\documentclass[11pt]{article}
\PassOptionsToPackage{dvipsnames}{xcolor}
\usepackage{xspace}
\usepackage{enumerate}
\usepackage{graphicx}
\usepackage{booktabs}
\usepackage{xcolor}
\usepackage{pifont}
\usepackage{pagecolor}
\usepackage{titlesec}
\usepackage{caption}
\usepackage{subcaption}
\usepackage[table]{xcolor}
\usepackage[most]{tcolorbox}
\tcbuselibrary{listings, skins, breakable}
\usepackage{listings}
\usepackage[percent]{overpic} 
\usepackage{hyperref}
\usepackage[capitalize]{cleveref}
\usepackage{tikz}
\usepackage{multirow}
\usepackage{dirtytalk}
\usepackage{tabularx}
\usepackage{tocloft}

\crefname{section}{Sec.}{Secs.}
\Crefname{section}{Section}{Sections}
\Crefname{table}{Table}{Tables}
\crefname{table}{Tab.}{Tabs.}

\newcommand{\xmark}{\ding{55}}

\usepackage[most]{tcolorbox}
\tcbuselibrary{listings, skins, breakable}
\usepackage{xcolor}
\usepackage{listings}

\definecolor{neoblue}{RGB}{179,217,242}
\definecolor{codegray}{gray}{0.15}
\definecolor{coderule}{RGB}{150,200,230}

\lstdefinestyle{neopy}{
  language=Python,
  basicstyle=\ttfamily\small,
  keywordstyle=\color{blue!70!black}\bfseries,
  stringstyle=\color{green!40!black},
  commentstyle=\color{black!60},
  numberstyle=\tiny\color{black!50},
  numbers=left,
  numbersep=8pt,
  showstringspaces=false,
  breaklines=true,
  tabsize=4,
  keepspaces=true,
}

\newtcblisting{neocodelst}[1]{
  listing only,
  listing options={style=neopy},
  colback=white,
  colframe=neoblue,
  enhanced,
  sharp corners,
  boxrule=0.8pt,
  left=8pt, right=8pt, top=6pt, bottom=6pt,
  breakable,
  title={#1},
  attach boxed title to top left = {yshift=-2mm, xshift=2mm},
  boxed title style = {colback=white, colframe=neoblue, boxrule=0.8pt},
  borderline west  = {3pt}{0pt}{neoblue!90},
}

\definecolor{tabbaseline}{rgb}{0.7, 0.85, 0.95} 
\definecolor{tabfirst}{rgb}{1, 0.7, 0.7} 
\definecolor{tabsecond}{rgb}{1, 0.85, 0.7} 
\definecolor{tabthird}{rgb}{1, 1, 0.7} 

\definecolor{rowblue}{RGB}{220,230,240}
\definecolor{myorchid}{RGB}{150,10,30}
\definecolor{myblue}{RGB}{10,30,250}
\definecolor{mygreen}{RGB}{10,120,10}

\newcommand{\redcross}{\textcolor{red!70!black}{\xmark}}
\newcommand{\greentick}{\textcolor{mygreen}{\textbf{\checkmark}}}

\newcommand{\disk}{{\mathtt D}}
\newcommand{\ram}{{\mathtt M}}
\newcommand{\cpu}{{\mathtt P}}
\newcommand{\llama}{Llama-3.2 3B}
\newcommand{\qwen}{Qwen-2.5 3B}
\newcommand{\llamasize}{5.98}
\newcommand{\qwensize}{6.32}
\newcommand{\green}[1]{\textcolor{ForestGreen}{#1}}

\newcommand{\red}[1]{\textcolor{BrickRed}{#1}}

\newcommand{\blue}[1]{\textcolor{RoyalBlue}{#1}}
\newtcolorbox[auto counter]{myfloatbox}[2][]{%
  enhanced,
  floatplacement=tb,
  float,
  colback=white,
  colframe=gray,
  arc=8pt,
  boxrule=1pt,
  title={Box~\thetcbcounter: #2},
  #1
}

\definecolor{systemgray}{RGB}{240,240,255}
\definecolor{userblue}{RGB}{230,245,255}
\definecolor{assistantgreen}{RGB}{230,255,240}
\definecolor{framegray}{RGB}{180,180,200}
\definecolor{frameblue}{RGB}{150,200,230}
\definecolor{framegreen}{RGB}{150,220,180}

\tcbset{
  system/.style={colback=systemgray, colframe=framegray, title=System Prompt, fonttitle=\bfseries},
  user/.style={colback=userblue, colframe=frameblue, title=User Query, fonttitle=\bfseries},
  assistant/.style={colback=assistantgreen, colframe=framegreen, title=Assistant Response, fonttitle=\bfseries}
}

\newtcolorbox[auto counter]{myfloatbox2}[2][]{%
  enhanced,
  floatplacement=tb,
  float,
  colback=white,
  colframe=black,
  boxrule=1pt,
  arc=8pt,
  title={Template~\thetcbcounter: #2},
  fonttitle=\bfseries,
  label={#1} 
}

\usepackage{amsmath,amsfonts,bm}

\usepackage[linesnumbered,ruled,vlined]{algorithm2e}

\SetKwComment{Comment}{\color{green!50!black}// }{}

\SetKwProg{Function}{function}{}{}

\newcommand{\tp}[1]{{#1}^{\top}}

\newcommand{\vc}[1]{\mathbf{#1}}

\newcommand{\calI}{{\cal I}}
\newcommand{\calJ}{{\cal J}}

\usepackage[margin=1in, top=1in]{geometry}
\usepackage{graphicx}
\usepackage{fancyhdr}
\usepackage{hyperref}
\usepackage{xcolor}
\usepackage{titlesec}
\usepackage{tcolorbox}
\usepackage{titling}
\tcbuselibrary{skins}

\usepackage[T1]{fontenc}
\usepackage{tgheros}
\usepackage[utf8]{inputenc}

\usepackage{amsmath,amssymb,bm}
\usepackage{newtxtext}
\usepackage{newtxmath} 

\AtBeginDocument{%

}

\providecommand{\titlefont}{\sffamily\bfseries}
\providecolor{qc_darkblue}{rgb}{0.008,0.063,0.247}
\providecolor{qc_blue}{rgb}{0.164,0.164,0.914}

\titleformat{\title}
{\titlefont\LARGE\color{qc_blue}}{}{0pt}{}

\titleformat{\section}
{\titlefont\Large\bfseries\color{qc_darkblue}}{\thesection}{1em}{}

\titlespacing*{\section}{0em}{1em}{.6em}

\newtcolorbox{titlebox}{
  enhanced,
  colback=white,
  boxrule=0pt,
  opacityback=0,
  opacityframe=0,
  width=0.95\textwidth,
  center
}

\makeatletter
\newcommand{\contactinfo}[1]{\def\@contactinfo{#1}}
\makeatother
\contactinfo{}

\fancypagestyle{titlepage}{
  \fancyhf{}
  \fancyfoot[L]{\footnotesize Qualcomm AI Research is an initiative of Qualcomm Technologies, Inc.}
  \fancyfoot[R]{\footnotesize\thepage}
  
}
\renewcommand\abstract{%
    \setlength{\parskip}{.8em}
    \par
}

\title{Paper Title}
\date{February 23, 2024}
\author{Author1, Author2}
\contactinfo{\{author1, author2\}@qualcomm.com}

\begin{document}
\thispagestyle{titlepage}

\begin{tikz}[remember picture,overlay]
  \node[anchor=north east, inner sep=0pt]
    at ([xshift=5cm,yshift=5.5cm]current page.north east)
    {\includegraphics[width=10cm,angle=-75,origin=c]{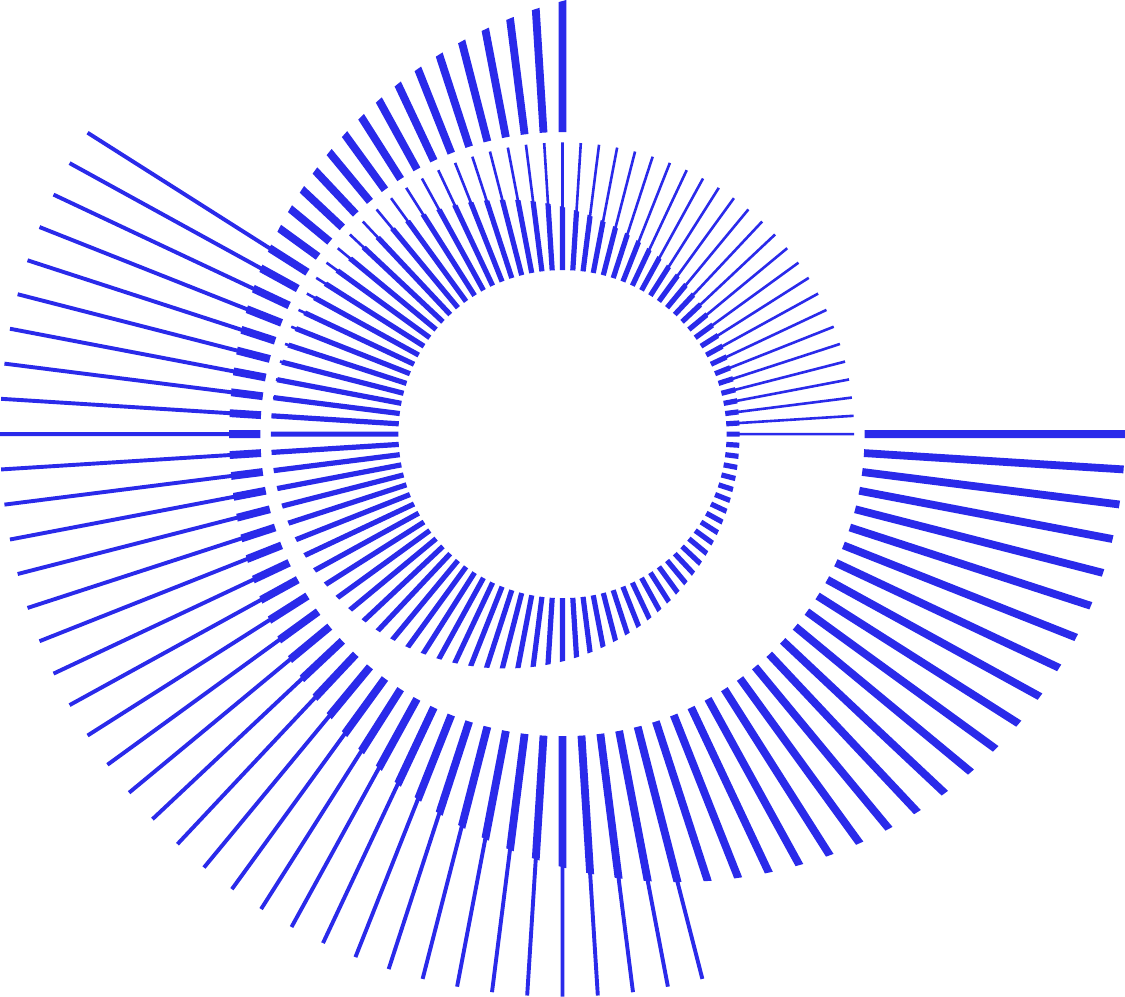}};
\end{tikz}

\begin{figure}[t]
    \vspace*{-1cm}
    \hspace*{-0.6cm} 
    \includegraphics[width=4.0cm]{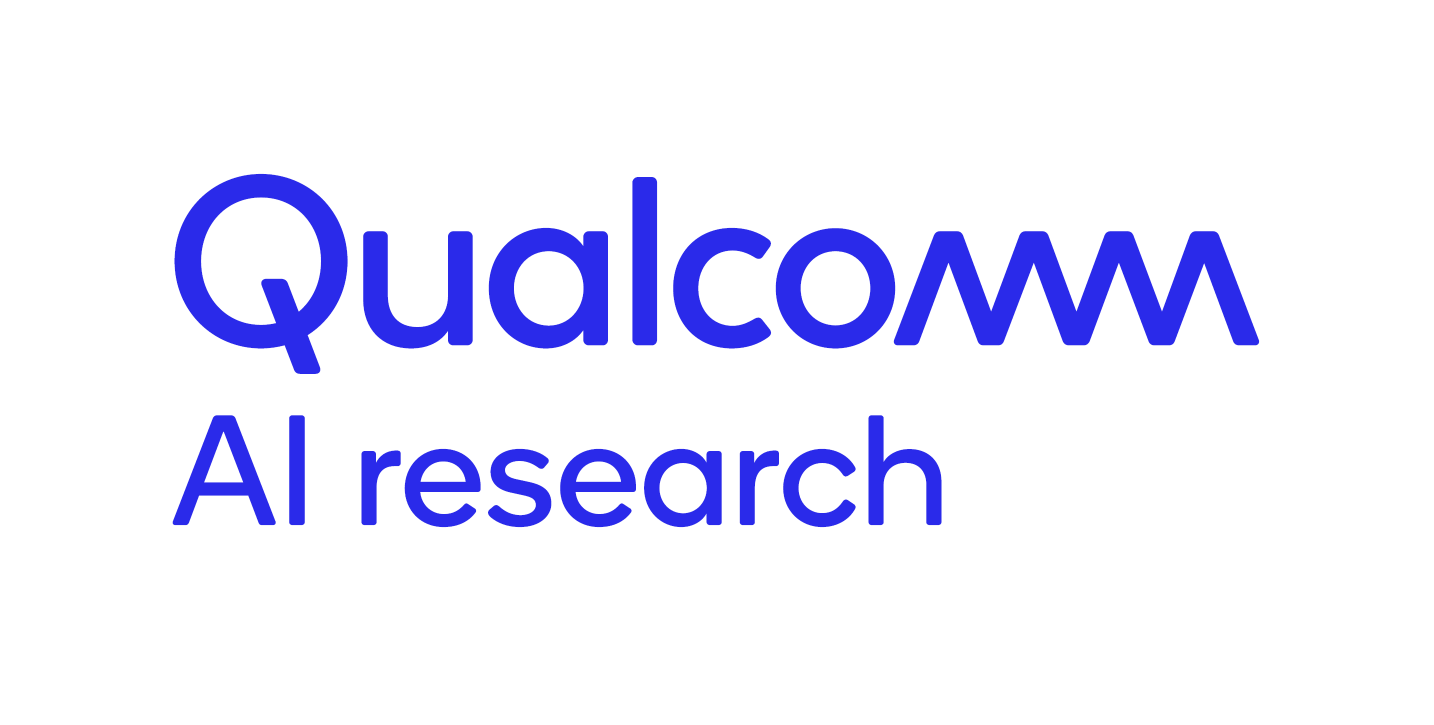}
    \vspace*{-0.5cm}
\end{figure}

\title{Techniques for Peak Memory Reduction for LoRA Fine-tuning of LLMs on Edge Devices}
\date{\today}
\author{
    Hassan Dbouk$^\dagger$,
    Matthias Reisser$^\dagger$,
    Prathamesh Mandke$^*$,
    Likhita Arun Navali,
    Christos Louizos
}
\contactinfo{}

\begin{titlebox}
{\titlefont\huge\bfseries\color{qc_darkblue}\thetitle}\\[1em]

\makeatletter
{\titlefont\color{qc_blue}\@author\par
 \ifx\@contactinfo\@empty
 \else
   \vspace{-.0em}
   {\bfseries\emphfont\@contactinfo}%
 \fi}
\makeatother

\vspace{.5em}

\begin{abstract}
Fine-tuning of Large Language Models (LLMs) using Low-Rank Adaptation (LoRA) on an end-user's data offers personalized experiences while keeping data private, but faces severe memory constraints on consumer hardware. Peak memory during fine-tuning often exceeds device limits, especially for models with billions of parameters and long-context training data. This paper introduces a suite of complementary techniques to reduce memory footprint without sacrificing model quality: (1) base model quantization with on-the-fly dequantization, (2) memory-efficient checkpointing combining selective activation caching and disk offloading, (3) softmax approximation using semantically relevant token subsets, and (4) logits masking. Experiments on \llama{} and \qwen{} demonstrate up to $26\times$ and $28\times$ reduction in peak memory, enabling fine-tuning on resource-constrained devices.
\end{abstract}

\vspace{1em}
$\dagger$ \emph{equal contribution}\\$*$ \emph{work done while at Qualcomm}
\end{titlebox}


\addtocontents{toc}{\protect\setcounter{tocdepth}{-1}}
\section{Introduction}
\begin{table}[htb]
\centering
\caption{Peak memory required for one forward and backward pass (batch size $1$, FP32 compute) for various LLMs measured on a NVIDIA A100 GPU across varying context lengths. Please see Section \ref{sec:experiments} for details.}
\label{tab:abl}
\vskip 0.15in
\begin{tabular}{c | l | r r r r r}
\toprule
\multirow{2.5}{*}{\textbf{Model}} & \multirow{2.5}{*}{\textbf{Method}} &
\multicolumn{5}{c}{\textbf{Measured Peak Memory [GB]}} \\
\cmidrule(lr{.5em}){3-7}
 & & 1024 & 2048 & 4096 & 8192 & 16384 \\
\midrule
\multirow{6}{*}{\rotatebox[origin=c]{90}{\llama{}}} & FP32 Baseline & 27.60 & 29.97 & 47.96 & OOM & OOM \\
 & \ + LoRA ($r=16$) & 19.10 & 26.20 & 40.41 & 68.83 & OOM\\
 & \ \ + Quantization & 8.80 & 14.68 & 27.48 & 53.07 & OOM\\
 & \ \ \ + Checkpointing &2.37 &3.33 &6.29  &12.21& 24.05\\
 & \ \ \ \ + SM Approx. & 1.30& 2.43&  5.41& 11.58& 23.79\\
 & \ \ \ \ \ + Logits Masking  & 0.70 & 1.02 & 1.59 & 3.24 & 6.95\\
\midrule
\multirow{6}{*}{\rotatebox[origin=c]{90}{\qwen{}}} & FP32 Baseline  & 27.11 & 35.25 & 57.87 & OOM & OOM\\
 & \ + LoRA ($r=16$) & 21.68 & 30.51 & 48.38 & OOM & OOM\\
 & \ \ + Quantization  & 11.21 & 19.46 & 36.03 & 69.17 & OOM\\
 & \ \ \ + Checkpointing & 2.06 & 3.79& 7.29& 14.27 & 28.24\\
 & \ \ \ \ + SM Approx. &  1.26& 3.03& 6.58& 13.79 & 27.98\\
 & \ \ \ \ \ + Logits Masking &  0.66&  1.01& 1.71 & 3.90 & 8.34\\
\bottomrule
\end{tabular}
\end{table}
Large Language Models (LLMs) have advanced rapidly, with increasingly powerful architectures emerging at an accelerating pace~\cite{grattafiori2024llama, jiang2023mistral, team2023gemini}. While training these foundation models remains computationally expensive, there is growing interest in on-device personalization for smartphones, IoT systems, and embedded platforms~\cite{liu2024mobilellm}, \emph{i.e.} on the \say{edge}. This trend is driven by privacy concerns, latency requirements, and the need for context-aware user experiences~\cite{slamanig2025llms, peng2024pocketllm}.

Personalizing LLMs typically relies on parameter-efficient fine-tuning methods such as Low-Rank Adaptation (LoRA)~\cite{hu2022lora}, which update a small subset of parameters while freezing the base model. LoRA reduces compute and memory, however, even models in the 3B--8B range impose substantial memory demands during fine-tuning, making deployment on edge devices difficult.

The bottleneck stems from the \emph{backward pass}, which requires storing forward-pass activations. Longer context windows exacerbate this issue. Without addressing this memory barrier, on-device adaptation, along with its benefits of privacy, low latency, and offline operation, remains impractical.

\paragraph{Contributions} We introduce techniques to reduce the peak memory footprint of LoRA fine-tuning of LLMs in resource-constrained environments. Our approach proposes base model quantization, checkpointing, logits masking, and softmax approximation. 
Table~\ref{tab:abl} shows how these complementary techniques can achieve substantial memory savings. For example, \llama{} with LoRA and a 2048-token context can be reduced from \textbf{26.20 GB} to as little as \textbf{1.02 GB}. Even very long-context examples become feasible, requiring just \textbf{6.95 GB} for \textbf{16384} tokens. Our contributions will unlock new use-cases for personalized experiences while maintaining data privacy.

\section{Background and Related Work}
\label{app:background-related-work}


\subsection{On-device LLM Fine-tuning}

Deploying and fine-tuning LLMs on resource-constrained devices has attracted significant attention. MobiLLM \cite{li2025mobillm} introduces a server-assisted side-tuning approach that enables LLM fine-tuning directly on mobile devices. Computationally intensive operations are offloaded to a server, while sensitive data remains on-device. This hybrid approach highlights the growing interest in personalization on edge devices despite resource limitations while sacrificing some privacy and latency due to its hybrid approach. PocketLLM \cite{peng2024pocketllm} leverages MeZO \cite{malladi2023fine}, a zeroth-order optimization method, to enable memory-efficient on-device fine-tuning of LLMs. It demonstrates practical performance on real mobile devices, however ZO gradients cannot replace backpropagation in all use-cases yet. MobileLLM \cite{liu2024mobilellm} designs sub-billion-parameter models optimized for on-device deployment, using architectural techniques such as deep-and-thin structures, embedding sharing, and grouped-query attention. While hardware-aware architectures hold great promises, they do not gain enough pre-training attention compared to general-purpose LLMs to fine-tune on. NNTrainer \cite{moon2022new} is a framework that enables on-device training through fine-grained execution-order analysis and proactive tensor swapping, achieving up to $20\times$ memory reduction. It demonstrates these techniques on \emph{small-scale} Transformer and Tacotron2 models, the latter being roughly 10--20 M parameters in size.\footnote{The evaluated Tacotron2 and Transformer models use $\sim$6 GB of memory at batch size 128, showing their smaller order of scale.} The on-demand materialization properties of NNTrainer relate closely to our offloading strategy, however their approach does not integrate easily with the PyTorch ecosystem, making it impracticable for developer-friendly deployment.

\subsection{Parameter-Efficient Fine-Tuning}

Parameter-Efficient Fine-Tuning (PEFT) methods adapt large pre-trained models with minimal computational and memory overhead. Low-Rank Adaptation (LoRA) \cite{hu2022lora} is among the most effective PEFT techniques. LoRA represents weight updates as low-rank decompositions, reducing trainable parameters significantly. Instead of updating full weight matrices, LoRA introduces pairs of rank decomposition matrices while keeping original weights frozen. For example, applied to a 175B parameter model, LoRA reduces trainable parameters by $10{,}000\times$ with minimal quality loss \cite{hu2022lora}.

Despite LoRA's efficiency, memory consumption during fine-tuning remains challenging on edge devices due to activation storage, optimizer states, and intermediate computations--especially with long context lengths. To address these limitations, several extensions have been proposed. QLoRA \cite{dettmers2023qlora} combines LoRA with 4-bit quantization of the base model weights, dramatically reducing memory footprint while maintaining accuracy. It introduces \emph{double quantization} and \emph{paged optimizers} to minimize memory overhead from optimizer states and activations, enabling fine-tuning of 65B-parameter models on a single GPU. HyC-LoRA \cite{wang2025hyclora} introduces a hybrid compression strategy that combines low-rank adaptation with structured pruning and quantization. By jointly optimizing rank and sparsity, HyC-LoRA achieves further memory savings without sacrificing convergence speed or model quality. LoRA-FA \cite{zhang2023lorafa} focuses on reducing activation memory during backpropagation. It employs \emph{forward activation approximation} techniques to avoid storing full activations, instead reconstructing them on-the-fly during gradient computation. This approach significantly lowers peak memory usage, making LoRA more practical for edge devices with strict memory constraints.

Our proposed techniques are orthogonal to these LoRA extensions and unlock further memory efficiencies.
\subsection{Memory Optimization Techniques}

Several techniques address high memory requirements in training deep neural networks. Gradient checkpointing \cite{chen2016training} trades computation for memory by storing fewer activations and recomputing during backpropagation, reducing memory by up to $10\times$ at approximately $33\%$ extra computation. In Sec.~\ref{app:grad-ckpt}, we show that offloading significantly reduces the peak memory of gradient checkpointing, at the cost of extra I/O. Zero-Offload \cite{ren2021zero} offloads optimizer states, gradients, and parameters to CPU memory, enabling large-scale training on limited GPUs; however, this requires sufficient CPU memory, which edge devices lack. MONeT \cite{shah2020memory} optimizes memory allocation via operator fusion, rematerialization, and memory planning to reduce peak usage. Adacc \cite{chen2025adacc} propose an adaptive memory optimization framework that combines layer-wise tensor compression with activation checkpointing and a mixed-integer scheduling policy to reduce GPU memory footprint during large-scale language model training. Their approach primarily targets large server-side training pipelines rather than on-device fine-tuning.

Our work differs by proposing complementary techniques tailored for on-device LoRA fine-tuning, including base model quantization, strategic checkpointing with memory offloading, logits masking, and softmax approximation. These methods are framework-agnostic and easily integrated into libraries like PyTorch.

\subsection{Efficient Softmax}

The computational and memory cost of softmax has motivated numerous approximations. Hierarchical softmax organizes vocabulary into a tree, reducing complexity from $O(V)$ to $O(\log V)$ \cite{morin2005hierarchical,mikolov2013efficient}. We don't rely on any explicit hierarchy or clustering, instead using the learned token similarity of the pre-trained LLM directly. Negative sampling considers only sampled negatives, reducing gradient computation \cite{mikolov2013distributed}, which we show to be sub-par to our similarity based selection for LLM fine-tuning. Adaptive softmax exploits Zipfian distribution by clustering words, minimizing expected computation \cite{grave2017efficient} but requiring training new parameters. Hardware-aware approximations replace expensive operations with low-cost alternatives, achieving up to $13\times$ area and $2\times$ energy savings \cite{geng2018hardware}. LUT-based approximations enable 8-bit quantized softmax with less than $1\%$ accuracy loss for attention-heavy architectures \cite{vasyltsov2021efficient}. Hardware-aware softmax modifications are orthogonal to our work, but promising for attention-heavy transformer models beyond the loss computation.

\section{Setup and Notation}
\label{app:setup-notation}

\paragraph{Notation} Scalars are denoted by lowercase letters (\textit{e.g.}, $s$, $b$, $r$). Vectors are denoted by bold lowercase letters (\textit{e.g.}, $\vc{x}$, $\vc{y}$). Matrices are denoted by bold uppercase letters (\textit{e.g.}, $\vc{W}$, $\vc{A}$, $\vc{B}$). Sets are denoted by calligraphic letters (\textit{e.g.}, $\mathcal{J}$, $\mathcal{V}$, $\mathcal{L}$). We use standard indexing notation: for a matrix $\vc{W}$, $\vc{W}[i,:]$ refers to the $i$-th row, and $\vc{W}[:,j]$ refers to the $j$-th column. For a vector $\vc{a}$, $\vc{a}[:\!k]$ denotes the first $k$ entries of $\vc{a}$.

\paragraph{Node Representation} We model a neural network as set of compute nodes $\mathcal{L} = \{v_1, v_2, \ldots, v_k\}$, where, for simplicity, we assume each node $v_i$ has one input tensor and one output tensor (though our techniques generalize to nodes with multiple inputs and outputs). We assume the last node $v_k$ computes the loss value used for backpropagation.

\paragraph{Compute Architecture} We assume a general-purpose processor $\cpu$ equipped with on-chip memory $\ram$ (\emph{e.g.}, RAM or cache), and access to off-chip storage $\disk$ (\emph{e.g.}, SSD or higher order cache) for reading and writing tensors. This design reflects typical edge devices where main memory is limited but disk storage is available. The processor supports a set of atomic operations for managing nodes and tensors during computation, summarized in Table~\ref{tab:atomic-ops}. Note that throughout we assume availability of FP32 matmul kernels only and consequently FP32 activations - although our methods generalize to arbitrary activation precision, changing the absolute peak memory measurement values correspondingly without changing conclusions.

\begin{table}[tb]
\centering
\caption{Atomic operations by the processor ($\cpu$)}
\label{tab:atomic-ops}
\footnotesize
\begin{tabularx}{\columnwidth}{l l X}
\toprule
\textbf{Operation} & \textbf{Params} & \textbf{Description} \\
\midrule
\texttt{load\_node} & $v$ & Load node $v$ from $\disk$ to $\ram$ \\
\texttt{load\_tensors} & $\{x_i\}$& Load tensors $\{x_i\}$ from $\disk$ to $\ram$ \\
\texttt{save\_tensors} & $\{x_i\}$ & Save tensors $\{x_i\}$ from $\ram$ to $\disk$ \\
\texttt{free\_tensors} & $\{x_i\}$ & Remove tensors $\{x_i\}$ from $\ram$ \\
\texttt{free\_node} & $v$ & Remove node $v$ from $\ram$ \\
\texttt{mark\_grad} & $x$ & Mark tensor $x$ for gradient computation \\
\texttt{fwd} & $v, x$ & Forward pass with autodiff enabled\\
\texttt{fwd\_no\_grad} & $v, x$ & Forward pass without autodiff \\
\texttt{grad} & $x$ & Get gradient of tensor $x$\\
\texttt{backward} & $x, g$ & Run backward pass starting from $x$ using output gradient $g$ if provided\\
\bottomrule
\end{tabularx}
\end{table}

\paragraph{Profiling} 
We use \llama{} \cite{grattafiori2024llama} and \qwen{} \cite{qwen2025qwen25technicalreport} in the main text. Unless otherwise stated, all peak-memory results measure a single forward--backward pass with batch size 1, FP32 compute, and LoRA adapters of rank $16$ on the query and value projections. Profiling is done on an NVIDIA A100 GPU with 80GB of VRAM. These off-target measurements are intended to isolate peak-memory behavior across methods; Section~\ref{ssec:on-device} reports deployment measurements on mobile hardware. Additional dataset, profiling, and hyperparameter details are in Appendix~\ref{app:setup}.

\section{Methodology} \label{sec:experiments}

\begin{algorithm}[thbp]
  \caption{On-the-fly Dequantization}
  \label{alg:dequant}
  \KwIn{Quantized weight matrix $\vc{W}_q$, quantization parameters $s, b$, input vector $\vc{x}$ \tcp{All in $\ram$}}
    
  \tcp{Forward pass}
  
  $\vc{W}_{\mathrm{fp32}} \gets s \cdot \vc{W}_q + b$ \tcp{Dequantize}
  
  \tcp{Compute base activations}
  $\vc{y}_{\text{base}} \gets \vc{W}_{\mathrm{fp32}} \cdot \vc{x}$

  \texttt{free\_tensors}$(\vc{W}_{\mathrm{fp32}})$
  
  \tcp{Complete forward, compute $d\vc{y}$}
  $\vc{W}_{\mathrm{fp32}} \gets s \cdot \vc{W}_q + b$ \tcp{Rematerialize}
  $d\vc{x} \gets \vc{W}_{\mathrm{fp32}}^\top \cdot d\vc{y}$
  
  \texttt{free\_tensors}$(\vc{W}_{\mathrm{fp32}})$

  \tcp{Complete backward}
  
\end{algorithm}

\subsection{Base Model Quantization}

Quantization is a key technique for compressing large language models and enabling deployment in resource-constrained environments, particularly in LoRA-based fine-tuning where base weights remain frozen and can be aggressively compressed. We adopt a standard post-training mixed-precision quantization strategy, compressing most linear layers (\(\vc{W}\)) to INT4 while retaining higher precision for quantization-sensitive components such as the output head (INT8) and input embeddings (INT16). This reduces the model footprint from \llamasize{} GB to 2.59 GB for \llama{}, significantly easing storage and bandwidth constraints. However, matmul with compressed weights and FP32 activations (which we assume throughout training) either requires specialized kernels or weight dequantization. To avoid memory inflation, we propose \textbf{on-the-fly dequantization}, where quantized weights are stored off-chip and re-materialized as \(\vc{W}_{\mathrm{fp32}}\) only when needed during forward and backward passes, then immediately discarded (Alg.~\ref{alg:dequant}). This eliminates the need to retain full-precision weights across passes, reducing peak memory by approximately the size of the base model.

\begin{table*}[tb]
\centering
\caption{Peak memory usage during one forward and backward pass for FP32 and quantized \llama{} and \qwen{} models with FP32 LoRA across varying context lengths. Both models are provided in BF16 on Hugging Face, necessitating upcasting to FP32.}
\label{tab:peak-mem-quant}
\vskip 0.15in
\resizebox{0.9\textwidth}{!}{%
\begin{tabular}{l | l| r | r r r r r r}
\toprule
\multirow{2.5}{*}{\textbf{Model}} & \multirow{2.5}{*}{\textbf{Matmul Precision}} & \multirow{2.5}{*}{\textbf{Size [GB]}} &
\multicolumn{6}{c}{\textbf{Measured Peak Memory [GB]}} \\
\cmidrule(lr{.5em}){4-9}
 &  &  & 512 & 1024 & 2048 & 4096 & 8192 & 12288 \\
\midrule
\multirow{3}{*}{\llama{}} & BF16 + FP32 (naive) & \llamasize{} & 15.55 & 19.10& 26.20& 40.41 & 68.83 & OOM  \\
 & BF16 + FP32 (Alg.~\ref{alg:dequant}) &\llamasize{}  & 11.29 & 14.36 & 20.95& 35.16 & 63.58 & OOM  \\
 & INT4 + FP32 (Alg.~\ref{alg:dequant})$\dagger$& 2.59 & 6.08 & 8.80 & 14.68 & 27.48 & 53.07  & OOM \\
\midrule
\multirow{3}{*}{\qwen{}} & BF16 + FP32 (naive) & \qwensize{} & 17.16 & 21.68 & 30.51 & 48.38 & OOM  & OOM \\
& BF16 + FP32 (Alg.~\ref{alg:dequant}) & \qwensize{} & 12.13 & 16.08  &  24.89 & 42.76  & OOM & OOM \\
 & INT4 + FP32 (Alg.~\ref{alg:dequant})$\dagger$  & 2.34 & 7.63 & 11.21 & 19.46 & 36.03 & 69.17  & OOM \\
\bottomrule
\multicolumn{9}{l}{\small $\dagger$Refer to the main text for exact quantization settings.} \\
\end{tabular}
\vspace{-0.5cm}
}
\end{table*}


\paragraph{Profiling} Despite this substantial reduction in parameter memory, profiling results (Table~\ref{tab:peak-mem-quant}) show that peak memory during training remains dominated by activations, which scale with context length and quickly lead to OOM beyond 8K tokens. Thus, while quantization effectively minimizes parameter storage and memory bandwidth, activation memory remains the primary bottleneck for long-context fine-tuning. Unless specified otherwise, all subsequent discussions use these quantized models.

\begin{algorithm}[!t]
  \caption{Memory-Efficient Checkpointing for LoRA Fine-tuning}
  \label{alg:checkpointing}
  \KwIn{Nodes $\mathcal{L} = (v_1,\dots,v_k)$, inputs $\vc{x}_0$}
  \KwOut{Loss $l$, gradients $\{\vc{g}_i\}_{i=0}^{k-1}$}

  \tcp{Stage I: Forward pass with caching (load from $\disk$ to $\ram$ as needed)}
  \texttt{load\_tensors}$(\vc{x}_0)$
  
  \For{$i \gets 1$ \KwTo $k-1$}{
    \texttt{load\_node}$(v_i)$
    
    $\vc{x}_i \gets \texttt{fwd\_no\_grad}(v_i, \vc{x}_{i-1})$ 
    
    \texttt{save\_tensors}$(\vc{x}_{i-1})$
    
    \texttt{free\_tensors}$(\vc{x}_{i-1})$
    
    \texttt{free\_node}$(v_i)$
  }

  \tcp{Stage II: Compute loss and start backprop}
  \texttt{load\_node}$(v_k)$
  
  \texttt{load\_tensors}$(\vc{x}_{k-1})$
  
  \texttt{mark\_grad}$(\vc{x}_{k-1})$
  
  $l \gets \texttt{fwd}(v_k, \vc{x}_{k-1})$
  
  \texttt{backward}$(l)$
  
  $\vc{g}_{k-1} \gets \texttt{grad}(\vc{x}_{k-1})$
  
  \texttt{free\_tensors}$(\vc{x}_{k-1})$
  
  \texttt{free\_node}$(v_k)$

  \tcp{Stage III: Backward pass with rematerialization}
  \For{$i \gets k-1$ \KwTo $1$}{
    \texttt{load\_node}$(v_i)$
    
    \texttt{load\_tensors}$(\vc{x}_{i-1})$
    
    \texttt{mark\_grad}$(\vc{x}_{i-1})$
    
    $\vc{x}_i \gets \texttt{fwd}(v_i, \vc{x}_{i-1})$
    
    \texttt{backward}$(\vc{x}_i, \vc{g}_i)$
    
    $\vc{g}_{i-1} \gets \texttt{grad}(\vc{x}_{i-1})$
    
    \texttt{free\_tensors}$(\vc{x}_{i-1}, \vc{x}_i)$
    
    \texttt{free\_node}$(v_i)$
    }
  \Return{$l$, $\{\vc{g}_i\}_{i=0}^{k-1}$}

\end{algorithm}

\subsection{Checkpointing}

To enable memory-efficient on-device fine-tuning, we propose a checkpointing strategy that jointly performs selective activation rematerialization and off-chip tensor offloading. We model the network as an ordered sequence of compute nodes $\mathcal{L}=(v_1,\dots,v_k)$, where node boundaries define which activations are cached versus recomputed during backpropagation. Unlike traditional checkpointing \cite{chen2016training} (Appendix~\ref{app:grad-ckpt}), which focuses solely on recomputation, our approach (Alg.~\ref{alg:checkpointing}) also dynamically loads and evicts weights and activations from off-chip storage, reducing peak on-chip memory. Specifically, we cache (off-chip) only selected boundary activations while rematerializing intermediate activations within each node, and load weights on demand, discarding them immediately after use. We instantiate this scheme using a canonical transformer decomposition into $k=n+2$ nodes (embedding layer, $n$ decoder layers, and output layer), with node placement guided by the memory budget and empirical peak-memory profiling. To minimize recomputation overhead, we greedily maximize the final node $v_k$, which does not require rematerialization; in practice, this node contains the LM head\footnote{We include logit, softmax, and loss calculations in the LM head}, which we observe to be the dominant memory bottleneck due to large vocabulary-dependent activations. 

\begin{table}[!htbp]
\centering
\caption{Peak memory evolution during a checkpointed forward and backward pass (Alg.~\ref{alg:checkpointing}) across context lengths for \llama{} and \qwen{}. For each context length, we \underline{underline} the first stage at which the final peak memory is reached. Across all context lengths, the logits and loss computation (\texttt{Stage II}) consistently emerges as the peak memory bottleneck. }
\label{tab:peak-mem-bottleneck}
\vskip 0.15in
\begin{tabular}{c | c| c | r r r r r r r}
\toprule
\multirow{2.5}{*}{\textbf{Model}} & \multirow{2.5}{*}{\textbf{Stage}} & \multirow{2.5}{*}{\textbf{Node}} &
\multicolumn{7}{c}{\textbf{Measured Peak Memory [GB]}} \\
\cmidrule(lr{.5em}){4-10} 
 &  &  & 512 & 1024 & 2048 & 4096 & 8192 & 12288 & 16384 \\
\midrule
\multirow{10}{*}{\rotatebox{90}{\llama{}}} & \multirow{4.5}{*}{\texttt{I}} & \blue{\texttt{embeddings}} & 0.01 & 0.02 & 0.05 & 0.09 & 0.19 & 0.28 & 0.38 \\
& & \green{\texttt{decoder[0]}} & 0.23 & 0.28 & 0.38 & 0.57 & 1.01 & 1.48 & 1.95 \\
& & \vdots & \multicolumn{7}{c}{\vdots} \\
& & \green{\texttt{decoder[27]}} & 0.23 & 0.28 & 0.38 & 0.57 & 1.01 & 1.48 & 1.95 \\
\cmidrule{2-10}
& \texttt{II} & \red{\texttt{lm head}} & \underline{2.11} & \underline{2.37} & \underline{3.33} & \underline{6.29} & \underline{12.21} & \underline{18.13} & \underline{24.05} \\
\cmidrule{2-10}
& \multirow{3.5}{*}{\texttt{III}} & \green{\texttt{decoder[27]}} & 2.11 & 2.37 & 3.33 & 6.29 & 12.21 & 18.13 & 24.05 \\
& & \vdots & \multicolumn{7}{c}{\vdots} \\
& & \green{\texttt{decoder[0]}} & 2.11 & 2.37 & 3.33 & 6.29 & 12.21 & 18.13 & 24.05 \\
\midrule
\midrule
\multirow{10}{*}{\rotatebox{90}{\qwen{}}} & \multirow{4.5}{*}{\texttt{I}} & \blue{\texttt{embeddings}} & 0.01 & 0.02 & 0.03 & 0.06 & 0.13 & 0.19 & 0.25 \\
& & \green{\texttt{decoder[0]}} & 0.24 & 0.29 & 0.40 & 0.65 & 1.21 & 1.78 & 2.35 \\
& & \vdots & \multicolumn{7}{c}{\vdots} \\
& & \green{\texttt{decoder[35]}} & 0.24 & 0.29 & 0.40 & 0.65 & 1.21 & 1.78 & 2.35 \\
\cmidrule{2-10}
& \texttt{II} & \red{\texttt{lm head}} & \underline{1.76} & \underline{2.06} & \underline{3.79} & \underline{7.28} & \underline{14.27} & \underline{21.26} & \underline{28.24} \\
\cmidrule{2-10}
& \multirow{3.5}{*}{\texttt{III}} & \green{\texttt{decoder[35]}} & 1.76 & 2.06 & 3.79 & 7.28 & 14.27 & 21.26 & 28.24 \\
& & \vdots & \multicolumn{7}{c}{\vdots} \\
& & \green{\texttt{decoder[0]}} & 1.76 & 2.06 & 3.79 & 7.28 & 14.27 & 21.26 & 28.24 \\
\bottomrule
\end{tabular}
\vspace{-0.1cm}
\end{table}

\paragraph{Profiling} Table~\ref{tab:peak-mem-bottleneck} reports the evolution of \llama{} and \qwen{} peak memory across nodes for varying context lengths. While checkpointing introduces additional compute (recomputing uncached activations), it significantly reduces memory: for example, at 8K tokens, \llama{} peak memory drops from 53.07\,GB (quantized only) to 12.21\,GB, and enables execution at longer contexts (12K–16K tokens) that otherwise cause OOM. Nevertheless, profiling shows that activation memory, particularly in the output layer, remains the primary bottleneck, motivating further optimization for long-context fine-tuning. Unless stated otherwise, all subsequent discussions use this checkpointing configuration.

\subsection{Softmax Approximation}\label{ssec:sm-approx}

\paragraph{Traditional Softmax} 
During on-device fine-tuning and after computing the last-layer hidden representations $\vc{h}_{1:T}$ from the input sequence $x_{1:T}$, we wish to compute the next-token probabilities of the target sequence $y_{1:T}$ for loss calculation. 
\begin{figure}[tbp]
    \centering
    \begin{subfigure}[b]{.8\textwidth}
        \centering
        \includegraphics[width=\textwidth,trim=0 1cm 0 1cm]{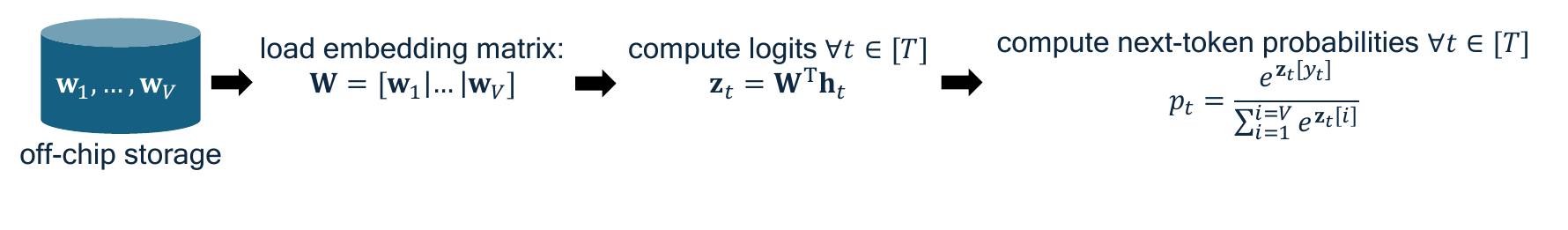}
        \caption{Traditional Full Softmax}
        \label{fig:full_sm}
    \end{subfigure}
    
    \vspace{1em} 
    \begin{subfigure}[b]{.8\textwidth}
        \centering
        \includegraphics[width=\textwidth,trim=0 0 0 0.5cm]{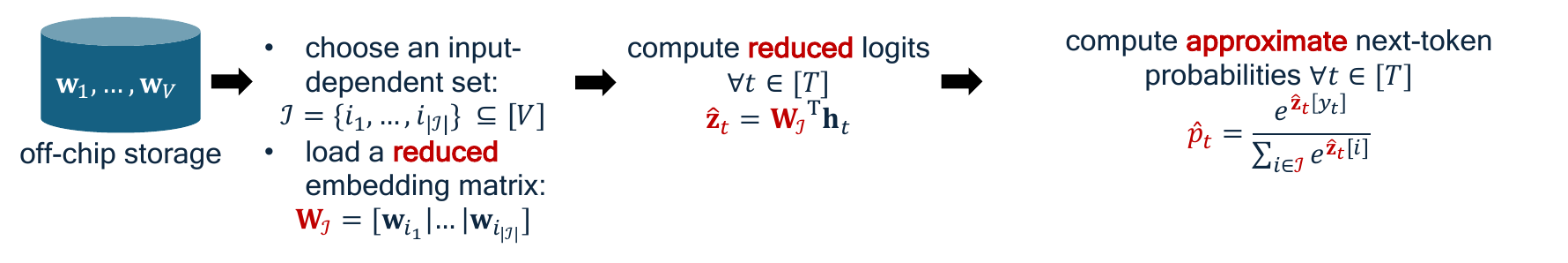}
        \caption{Reduced Vocabulary Softmax Approximation}
        \label{fig:approx_sm}
    \end{subfigure}
    
    \caption{Illustration of the traditional (a) and proposed (b) softmax computation.}
    \label{fig:full_vs_approx_sm}
\end{figure}
As illustrated in Fig~\ref{fig:full_sm}, the traditional approach would use all $V$ word embeddings to compute logits $\vc{z}_t$, from which we compute $\forall t$:
\begin{equation}
    p_t = \frac{e^{\vc{z}_t[y_t]}}{\sum_{i=1}^{V}e^{\vc{z}_t[i]}}
\end{equation}
The denominator term in the softmax sums over all the logits in the vocabulary, which can be as large as $128$k (Llama-3) or $256$k (Gemma-2) for modern LLMs. Intuitively, maximizing $p_t$ can be understood as maximizing the score of the correct token $y_t$ while minimizing the score of all remaining tokens. Geometrically, this corresponds to “pushing” the token representation $\vc{h}_t$ closer to $\vc{w}_{y_t}$ and away from all other word embeddings. 

\paragraph{Intuition for Softmax Approximation}
Pretrained models have already learned to understand the context and relationships between tokens. Tokenizers are designed to handle a wide range of text, including common words, symbols, and rare terms. For pretrained models, next-token probability vectors end up being very sparse. Furthermore, in specific fine-tuning use-cases, not all tokens in the vocabulary matter as many of them can be semantically irrelevant to the specific context. Instead, we wish to distinguish between tokens close to the decision boundary. 

\paragraph{Reducing the Vocabulary Size}
Leveraging this intuition, we propose to reduce the size of the \say{effective} vocabulary used during softmax computation, illustrated in Fig~\ref{fig:approx_sm}: assume we have an \say{oracle} that determines a good subset of entries $\calJ \subseteq [V]$, then we can afford to load a reduced embedding matrix based on $\calJ$ into memory, from which we compute a logits vector and probabilities that are reduced in length from $V$ to $|\calJ|$. Doing so allows us to:
\begin{itemize}
    \item reduce off-chip $\iff$ on-chip I/O, since we load $|\calJ| \leq V$ rows,
    \item reduce the peak on-chip memory required due to loading a smaller word embedding matrix,
    \item reduce the peak on-chip memory required due to smaller logits sizes, which are used for softmax computation, loss calculation, and in both forward and backward passes,
    \item reduce compute of the LM head.
\end{itemize}

\paragraph{Choosing the Reduced Vocabulary Set}
The goal is to choose the smallest possible subset $\calJ \subseteq [V]$ such that the model converges to a desired quality. It is well understood in the language modeling literature \cite{mikolov2013efficient} that word embedding vectors capture the underlying tokens’ semantics geometrically. That is, tokens that are semantically similar will often have \say{closer} embedding vector representations. We leverage the fact that the embedding matrix $\vc{W}$ is frozen for LoRA fine-tuning and pre-compute $\calI(y_t,k)$: the top-$k$ \emph{closest} tokens to the target token $y_t$ in the sequence.  We first pre-compute the 
the embeddings' (or tokens') cosine similarity matrix $\vc{C}=\tp{\tilde{\vc{W}}}\tilde{\vc{W}}$ where $\tilde{\vc{W}}$ is the column-normalized embeddings matrix. We then sort $\vc{C}[i,:]$, \emph{i.e.} row $i$ of $\vc{C}$, in descending order, and collect the corresponding sorted indices $\vc{a}_i=[a_{i,1},\cdots,a_{i,V}]$ such that $\vc{C}[i,a_{i,1} ]\geq \cdots \geq \vc{C}[i,a_{i,V}]$.

Finally, we aggregate these $\vc{a}_i$ in matrix format $\vc{A}=[\vc{a}_1, \cdots ,\vc{a}_V ]$, to be used as a lookup table during fine-tuning. Specifically, for a certain $k$, we only need to store the first $k$ indices within $\vc{A}$ in INT32. For \llama, this implies $128256 \times k\times4=0.489 k$ MB of additional off-chip storage. Later, we show that $k=500$ offers a good trade-off in approximation quality for the XSum dataset \cite{xsum}. At this setting, the off-chip storage overhead is approximately $245$ MB. Note that $\vc{A}$ is use-case agnostic, and can be re-used across all fine-tuning applications. 

While the intuition behind token similarity and approximation of the softmax applies on a per-token position basis, we nonetheless must consider that during fine-tuning, the LLM processes an entire sequence of tokens in one-shot, and thus fetching a different set of word embeddings based on $\calI(y_t,k)$ for each token position is impractical. During fine-tuning, we construct $\calJ$ for a sequence by taking the union of all\footnote{In practice, we only include trainable tokens} target tokens' top-$k$ indices:
\begin{equation}
    \calJ = \cup_{t=1}^T  \calI(y_t,k) = \cup_{t=1}^T \vc{a}_{y_t}[:\!k]
\end{equation}
Note that, by construction, the first token $a_{i,1}$ will always be $i$, which guarantees for $k\geq 1$ that $\calJ$ contains the target tokens $y_{1:T}$.

\begin{table}[thbp]
\centering
\caption{Hyperparameters for LoRA fine-tuning on XSum}
\label{tab:hparams}
\begin{tabular}{l l}
\toprule
\textbf{Hyperparameter} & \textbf{Value} \\
\midrule
LoRA layers & q, v \\
LoRA rank ($r$) & 16 \\
LoRA alpha ($\alpha$) & 16 \\
Learning rate & 5e-4 \\
Batch size & 8 \\
Weight decay & 0.01 \\
Optimizer & AdamW \\
Training steps & 500 \\
Max sequence length & 2048 \\
\bottomrule
\end{tabular}
\vspace{-0.4cm}
\end{table}

\paragraph{Fine-tuning with Softmax Approximation}\label{sssec:sm-exp}
To assess the efficacy of our proposed softmax approximation strategy, we fine-tune \llama{} and \qwen{} on the XSum dataset \cite{xsum}, prepared under constraints for on-device adaptation (see Appendix~\ref{app:xsum} for details). The same experiments with SQuAD \cite{squad} can be found in Appendix~\ref{app:SQuAD}; additional datasets, model configs, and quantitative results are in Appendices~\ref{app:extra-configs} and~\ref{app:quantitative-results}. Specifically, we compare three methods: (1) full softmax over the entire vocabulary, (2) top-$k$ softmax using semantically similar tokens, and (3) rand-$k$ softmax with randomly sampled negatives. These experiments evaluate memory efficiency and convergence behavior. All models are fine-tuned using LoRA adaptation with identical hyperparameters (\emph{c.f.}  Table~\ref{tab:hparams}) ensuring fair comparison.

\begin{figure}[t]
    \centering
    \includegraphics[width=\columnwidth]{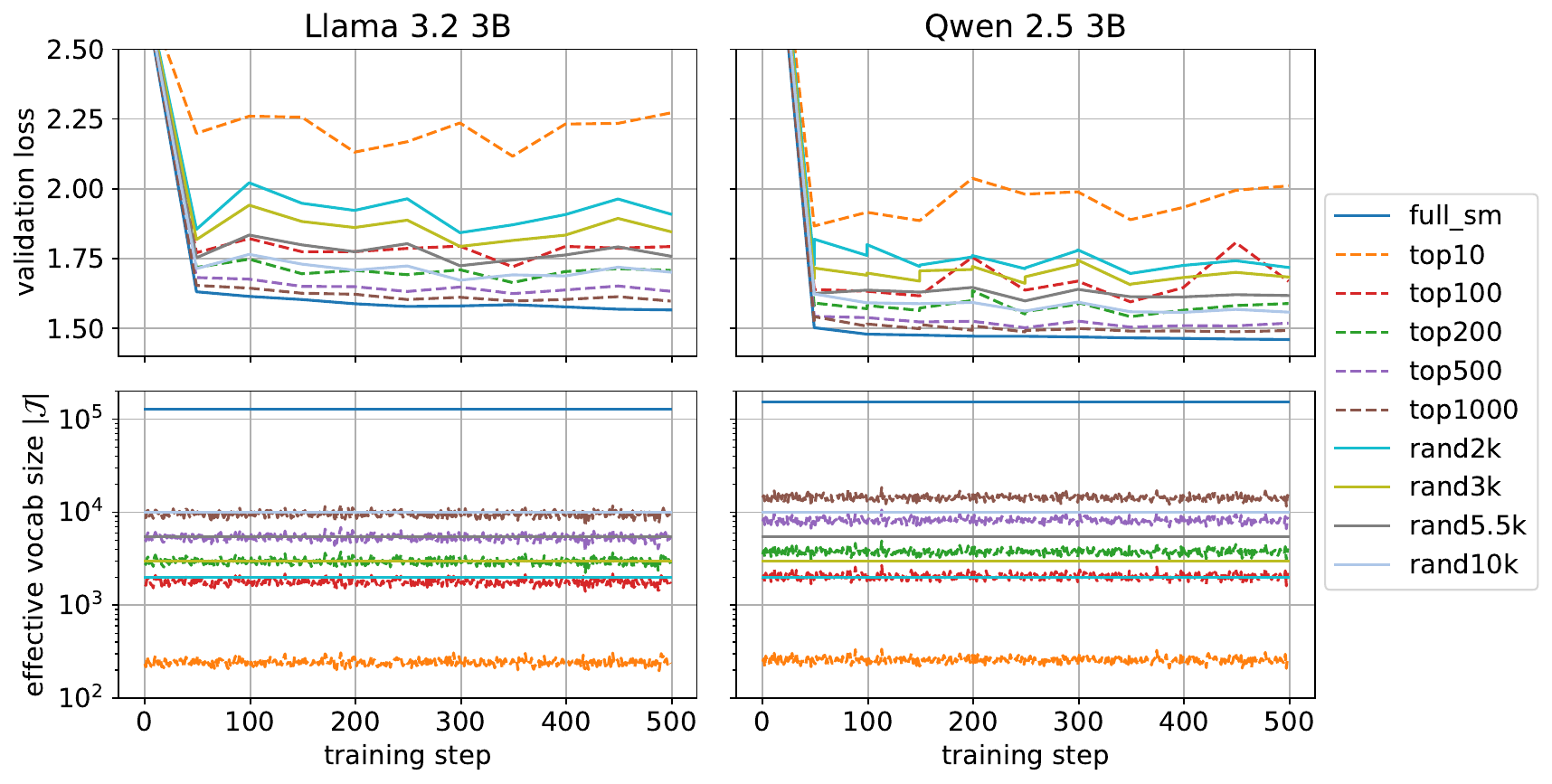}
    \caption{Validation loss vs. training steps and effective vocabulary size $|\mathcal{J}|$ for \llama{} and \qwen{} fine-tuned on XSum using different softmax methods. Top-$k$ maintains convergence quality comparable to full softmax while using a fraction of the vocabulary.}
    \label{fig:xsum_llama3}
\end{figure}

Figure~\ref{fig:xsum_llama3} shows validation loss and effective vocabulary size $|\mathcal{J}|$ across training steps. Top-$k$ softmax achieves convergence comparable to full softmax while using a fraction of the vocabulary. In contrast, rand-$k$ converges to worse solutions, underscoring the importance of semantic token selection (Appendix~\ref{app:sm-ablation}). The effective vocabulary for top-$k$ stays far smaller than the full set throughout training, yielding substantial memory savings (Table~\ref{tab:sm_approx}). The \qwen{} curves exhibit the same pattern.

\begin{table}[!t]
\centering
\caption{Peak memory vs.\ context length under checkpointing \& softmax approx. Levels for \llama{} and \qwen{}. Colors indicate memory bottleneck: \green{decoder} \& \red{output}.}
\label{tab:sm_approx}
\vskip 0.15in
\begin{tabular}{c | r | r r r r}
\toprule
\multirow{2.5}{*}{\textbf{Model}} & \multirow{2.5}{*}{\textbf{Context Length}} &
\multicolumn{4}{c}{\textbf{Measured Peak Memory [GB]}} \\
\cmidrule(lr{.5em}){3-6}
& & Full SM & Top-100 & Top-500 & Top-1000 \\
\midrule
\multirow{7}{*}{\rotatebox{90}{\llama{}}} & 512 & \red{2.11} & \green{0.37} & \red{0.51} & \red{0.79} \\
& 1024 & \red{2.37} & \green{0.54} & \red{0.88} & \red{1.30} \\
& 2048 & \red{3.33} & \green{0.86} & \red{1.77} & \red{2.43} \\
& 4096 & \red{6.29} & \red{2.31} & \red{4.20} & \red{5.41} \\
& 8192 & \red{12.21} & \red{6.26} & \red{9.86} & \red{11.58} \\
& 12288 & \red{18.13} & \red{10.83} & \red{15.80} & \red{17.74} \\
& 16384 & \red{24.05} & \red{15.64} & \red{21.72} & \red{23.79} \\
\midrule
\multirow{7}{*}{\rotatebox{90}{\qwen{}}} & 512 & \red{1.76} & \green{0.36} & \red{0.51} & \red{0.77} \\
& 1024 & \red{2.06} & \green{0.51} & \red{0.88} & \red{1.26} \\
& 2048 & \red{3.79} & \red{0.93} & \red{2.30} & \red{3.03} \\
& 4096 & \red{7.29} & \red{2.59} & \red{5.42} & \red{6.58} \\
& 8192 & \red{14.27} & \red{7.01} & \red{12.28} & \red{13.79} \\
& 12288 & \red{21.26} & \red{12.28} & \red{19.28} & \red{20.94} \\
& 16384 & \red{28.24} & \red{17.72} & \red{26.17} & \red{27.98} \\
\bottomrule
\end{tabular}
\end{table}

\paragraph{Profiling} 
Table~\ref{tab:sm_approx} illustrates the combined effect of quantization, checkpointing, and softmax approximation across different approximation levels. When applied on top of checkpointing, softmax approximation delivers substantial reductions in peak memory. For example, at 8K tokens, \llama{} drops from 12.21\,GB (checkpointing only) to 6.26\,GB with Top-100 approximation - a $1.9\times$ reduction. At shorter contexts (\textit{e.g.}, 512 tokens), the output layer ceases to be the bottleneck entirely under Top-100 approximation, shifting the peak memory to the decoder layers.

Even at extreme sequence lengths (12K tokens), where memory pressure is highest, Top-100 approximation reduces \llama{} from 18.13\,GB to 10.83\,GB, enabling configurations that would otherwise be infeasible on commodity hardware. The \qwen{} results show the same trend, confirming that softmax approximation, combined with checkpointing and quantization, is effective beyond the Llama setting.

Memory profiling of softmax approximation is token-dependent. Table \ref{tab:sm_approx} was obtained by profiling appropriately concatenated examples from the XSum dataset and setting all tokens to be trainable. These values therefore represent an upper-bound, while in practice, memory reduction depends on how many trainable tokens are present in the sequence. As visualized in Fig.~\ref{fig:xsum_llama3}, where tokens are masked correctly (Table \ref{tab:xsum-squad-stats}), $|\mathcal{J}|$ is fluctuating slightly across data-points.

\subsection{Logits Masking}\label{ssec:logits-masking}

\paragraph{Instruction Fine-tuning}
Instruction fine-tuning trains LLMs with annotated examples of desired behaviors and responses. 
Each sample (Box~\ref{box:xsum-example}) consists of non-trainable tokens (\textcolor{RedOrange}{system prompt} and \textcolor{blue}{relevant context}) and trainable tokens (\textcolor{green!60!black}{response}), where we use the term \say{token} loosely to also refer to a \say{word}.

\begin{myfloatbox}[label={box:xsum-example}]{An Instruction Fine-tuning Example from XSum \cite{xsum}}
\textcolor{orange}{System: You are a writing assistant that helps with summarizing text. You write summaries based on a user-provided text or article.}

\textcolor{blue}{Article: Officers searched properties in the Waterfront Park and Colonsay View areas of the city on Wednesday. Detectives said three firearms, ammunition and a five-figure sum of money were recovered. A 26-year-old man who was arrested and charged appeared at Edinburgh Sheriff Court on Thursday.}

\textcolor{green!60!black}{Summary: A man has appeared in court after firearms, ammunition and cash were seized by police in Edinburgh.}

\end{myfloatbox}

\paragraph{From Loss Masking to Logits Masking}
In instruction fine-tuning, it is standard practice \cite{huerta2024instruction} to \say{mask} non-trainable tokens during loss computation, as these correspond to tokens we do not want the model to generate at test time. Formally, if $x_{1:T}$ and $y_{1:T}$ are the input and target sequences, let $\mathcal{T} \subseteq [T]$ denote the set of trainable tokens. The masked causal language modeling loss is:
\begin{equation}
    \text{loss} = - \sum_{t=1}^T \mathbf{1}_{t\in\mathcal{T}} \log p_t = - \sum_{t\in\mathcal{T}} \log p_t
\end{equation}
where $p_t$ is the probability of predicting token $y_t$.

Loss masking still computes logits and softmax probabilities for all tokens, including non-trainable ones. We propose \textit{logits masking}: skip computing logits for non-trainable tokens entirely. This yields identical losses and gradients while reducing peak memory during the LM head computation. Effectively, it shortens the context length processed by the LM head in both, forward and backward pass.

\paragraph{Limitations}\label{sssec:limits-logits-masking}
Logits masking has two limitations. First, it applies only to fine-tuning settings that contain non-trainable tokens, and memory savings depend on the proportion of trainable tokens; if most tokens are trainable, savings are negligible. Second, its data dependency complicates batching for the LM head - assuming that other constraints allow micro-batches of size larger than one. This can be circumvented by processing the LM head in a loop over batch elements.

\begin{table}[!t]
\centering
\caption{Impact of logits masking on peak memory vs. context length at different percentages of trainable tokens; results include checkpointing and top-1000 softmax approximation. Colors indicate the memory bottleneck: \green{decoder} and \red{output}.}
\label{tab:logits-mask}
\vskip 0.15in
\begin{tabular}{c | r | r r r r  r r r r r r}
\toprule
\multirow{2.5}{*}{\textbf{Model}} & \multirow{2.5}{*}{\shortstack{\textbf{Context}\\\textbf{Length}}} &
\multicolumn{10}{c}{\textbf{Measured Peak Memory [GB]}} \\
\cmidrule(lr{1em}){3-12}
& & 10\% & 20\% & 30\% & 40\% & 50\% & 60\% & 70\% & 80\% & 90\% & 100\% \\
\midrule
\multirow{7}{*}{\rotatebox{90}{\llama{}}}& 512 & \green{0.37} & \red{0.37} & \red{0.47} & \red{0.51} & \red{0.56} & \red{0.61} & \red{0.64} & \red{0.71} & \red{0.76} & \red{0.79} \\
& 1024 & \green{0.54} & \red{0.60} & \red{0.70} & \red{0.83} & \red{0.91} & \red{1.02} & \red{1.09} & \red{1.16} & \red{1.23} & \red{1.30} \\
& 2048 & \green{0.86} & \green{0.86} & \red{1.02} & \red{1.21} & \red{1.39} & \red{1.59} & \red{1.72} & \red{1.91} & \red{2.16} & \red{2.43} \\
& 4096 & \green{1.55} & \green{1.55} & \red{1.59} & \red{1.92} & \red{2.43} & \red{3.04} & \red{3.63} & \red{4.24} & \red{4.84} & \red{5.41} \\
& 8192 & \green{3.01} & \green{3.01} & \red{3.24} & \red{4.44} & \red{5.62} & \red{6.81} & \red{7.98} & \red{9.16} & \red{10.38} & \red{11.58} \\
& 12288 & \green{4.47} & \green{4.47} & \red{5.08} & \red{6.89} & \red{8.77} & \red{10.59} & \red{12.37} & \red{14.16} & \red{15.94} & \red{17.74} \\
& 16384 & \green{5.93} & \green{5.93} & \red{6.95} & \red{9.42} & \red{11.82} & \red{14.26} & \red{16.65} & \red{19.04} & \red{21.40} & \red{23.79} \\
\midrule
\multirow{7}{*}{\rotatebox{90}{\qwen{}}} & 512 & \green{0.36} & \red{0.38} & \red{0.48} & \red{0.52} & \red{0.57} & \red{0.62} & \red{0.66} & \red{0.70} & \red{0.75} & \red{0.77} \\
& 1024 & \green{0.51} & \red{0.54} & \red{0.66} & \red{0.78} & \red{0.86} & \red{0.98} & \red{1.05} & \red{1.13} & \red{1.20} & \red{1.26} \\
& 2048 & \green{0.84} & \green{0.84} & \red{1.01} & \red{1.26} & \red{1.42} & \red{1.73} & \red{2.06} & \red{2.40} & \red{2.71} & \red{3.03} \\
& 4096 & \green{1.59} & \green{1.58} & \red{1.71} & \red{2.37} & \red{3.05} & \red{3.75} & \red{4.49} & \red{5.19} & \red{5.91} & \red{6.58} \\
& 8192 & \green{3.07} & \green{3.07} & \red{3.90} & \red{5.39} & \red{6.82} & \red{8.22} & \red{9.60} & \red{10.98} & \red{12.41} & \red{13.79} \\
& 12288 & \green{4.55} & \green{4.55} & \red{6.15} & \red{8.29} & \red{10.44} & \red{12.56} & \red{14.67} & \red{16.76} & \red{18.85} & \red{20.94} \\
& 16384 & \green{6.03} & \green{6.03} & \red{8.34} & \red{11.15} & \red{14.01} & \red{16.83} & \red{19.62} & \red{22.41} & \red{25.19} & \red{27.98} \\
\bottomrule
\end{tabular}
\end{table}

\paragraph{Profiling} When combined with quantization, checkpointing, and top-1000 softmax approximation, logits masking delivers significant memory savings without impacting performance for tasks where most tokens are fixed. Table~\ref{tab:logits-mask} summarizes peak memory across varying percentages of trainable tokens. As the proportion of trainable tokens decreases, peak memory consistently drops, until the bottleneck shifts from the \red{output} layer to the \green{decoder} layers. For example, in \llama{}, assuming 30\% trainable tokens reduces peak memory by approximately $1.7\times$-$3.5\times$ compared to the full-token case, while maintaining task accuracy. At extreme sequence lengths (16K tokens), memory drops from 23.79\,GB (100\% trainable) to 6.95\,GB at 30\% trainable tokens, enabling configurations that would otherwise be infeasible.

These gains are dataset- and task-dependent. For instance, on XSum, we observe an average of only 5\% trainable tokens per sample (see Table ~\ref{tab:xsum-squad-stats} in Appendix~\ref{app:xsum}), suggesting that logits masking can unlock substantial memory savings in realistic fine-tuning scenarios. In Appendix~\ref{app:lm_no_sm}, we present results of logits masking without SM approximation, demonstrating that the two techniques are complementary. Their combination leads to greater savings than either method alone.

\subsection{Overall Impact} Combining all four techniques yields dramatic memory savings. At $4$K tokens, the FP32 LoRA baseline for \llama{} requires 40.41 GB (Table~\ref{tab:peak-mem-quant}), while our optimized configuration with INT4 quantization, checkpointing, Top-1000 softmax approximation, and logits masking (30\% trainable tokens) reduces peak memory to just \textbf{1.59 GB}, a reduction of over $25\times$. For 16K tokens, FP32 baselines are completely infeasible (projected to exceed 80 GB)\footnote{The A100 GPU has 80GB of VRAM.}. In contrast, our optimized pipeline enables fine-tuning at this long context length with only \textbf{6.95 GB}, which is in the range of available memory of edge-devices such as phones and laptops. For \qwen{}, the FP32 baseline at 4K tokens consumes 48.38 GB, whereas the same optimized setup achieves \textbf{1.71 GB}, a $28\times$ improvement. Appendix~\ref{app:llama70b} reports analogous reductions for Llama-3.3 70B.

\subsection{On-Device Measurements}\label{ssec:on-device}

\begin{table}[!tb]
\centering
\caption{Latency and peak memory of \llama{} on a top-tier smartphone from 2025 with 12\,GB RAM under different combinations of checkpointing (Ckpt), softmax approximation (Top-$k$), and logits masking (Mask). Latency and memory are reported relative to the 32-token unoptimized baseline. The two ``All Methods'' rows demonstrate the combined effect at short and long contexts.}
\label{tab:on-device}
\vskip 0.15in
\begin{tabular}{l c l c r r}
\toprule
\textbf{Configuration} & \textbf{Ckpt} & \textbf{Top-$k$} & \textbf{Mask} & \textbf{Latency} & \textbf{Memory} \\
\midrule
Baseline (32 Tokens)      & \redcross & \redcross{} (full)   & \redcross & 100.00\% & 100.00\% \\
+Logits Masking           & \redcross & \redcross{} (full)   & \greentick & 98.77\%  & 100.00\% \\
+Softmax Approx           & \redcross & \greentick{} ($k$=200)  & \redcross & 93.89\%  & 42.80\% \\
+Logits \& SM Approx      & \redcross & \greentick{} ($k$=200)  & \greentick & 93.61\%  & 42.80\% \\
+Checkpointing            & \greentick & \redcross{} (full)   & \redcross & 118.42\% & 65.68\% \\
+Ckpt+Masking             & \greentick & \redcross{} (full)   & \greentick & 120.15\% & 65.68\% \\
+Ckpt+Softmax             & \greentick & \greentick{} ($k$=200)  & \redcross & 116.15\% & 7.38\% \\
\textbf{All Methods (32 Tokens)}   & \greentick & \textbf{\greentick{} ($k$=200)} & \greentick & \textbf{115.10\%} & \textbf{7.38\%} \\
\textbf{All Methods (2048 Tokens)} & \greentick & \textbf{\greentick{} ($k$=200)} & \greentick & \textbf{802.49\%} & \textbf{19.37\%} \\
\bottomrule
\end{tabular}
\end{table}

Thus far, all reported results were measured off-target on an NVIDIA A100 GPU using CUDA. To demonstrate the practicality of our techniques, we deploy the optimized pipeline on a top-tier smartphone from 2025 with 12 GB RAM. Table~\ref{tab:on-device} reports relative latency and peak memory for a single forward and backward pass under different combinations of our techniques. With all techniques combined and $k=200$, peak memory drops to \textbf{7.38\%} of the unoptimized baseline at 32 tokens, and to \textbf{19.37\%} at 2048 tokens, confirming that the memory reductions observed on the A100 transfer to commodity mobile hardware. The latency cost reflects the recomputation overhead of checkpointing and grows with context length, as expected.

\section{Discussion}
This work combines a set of orthogonal techniques - quantization, memory-efficient checkpointing, softmax approximation and logits masking — that collectively enable LoRA-based fine-tuning of LLMs on resource-constrained devices. While some of these methods are individually known, their integration into a cohesive pipeline tailored for on-device learning is both novel and impactful. The strength of our work lies in our memory-bottleneck driven approach and the synergy between these techniques, which in combination unlock long-context personalization on commodity mobile hardware. Moreover, this paper serves as a practical guide for practitioners seeking to understand and navigate the memory limitations of on-device fine-tuning, offering actionable insights grounded in empirical profiling and optimization.

\section{Limitations}
 A limitation of our evaluation is that the softmax approximation is validated only for text-only fine-tuning tasks. While quantization, checkpointing, and logits masking are not inherently tied to the data modality, our reduced-vocabulary softmax relies on token-level semantic similarity in the language-model embedding space, and its effectiveness may not directly generalize to multimodal, speech, retrieval-augmented, or other non-standard output spaces. The benefits of logits masking are also task-dependent: they are largest when only a small fraction of the sequence contains trainable target tokens, as in instruction fine-tuning, and diminish when most tokens contribute to the loss. Finally, the on-device profiling results depend on a specific mobile deployment stack and hardware configuration, so they should be interpreted as evidence of feasibility rather than a fully reproducible benchmark across edge devices.

\bibliographystyle{alpha}
\bibliography{refs}

\clearpage
\appendix
\addtocontents{toc}{\protect\setcounter{tocdepth}{2}}
\begingroup
\renewcommand{\contentsname}{Appendix Contents}
\setcounter{tocdepth}{2}
\tableofcontents
\endgroup

\clearpage
\section{Experimental Setup}\label{app:setup}
We conduct all training and profiling experiment on a single NVIDIA A100 GPU with 80GB of VRAM.

\subsection{Profiling}\label{app:profiling}
For all profiling measurements, we feed the LLMs pre-tokenized samples from the XSum dataset. To achieve a target context length, we concatenate examples until the desired length is reached, then chunk the sequence accordingly. We remove the \texttt{-100} masks from the labels so that all tokens are trainable (an overestimate), except for the logits-masking experiments where we sweep by masking the first $X\%$ of tokens. We use a batch size of 1.

\subsection{Dataset Preparation}\label{app:dataset}
For all datasets, we utilize their publicly available implementations from HuggingFace. For each model, we format every data point using its corresponding tokenizer's chat template with a maximum block size of 2048.
\subsubsection{XSum Dataset}\label{app:xsum}
The XSum \cite{xsum} dataset was prepared with a system prompt that instructs the model to act as a writing assistant focused on summarization, as shown in Box~\ref{box:xsum-example}.

Table~\ref{tab:xsum-squad-stats} presents key statistics about the XSum dataset used in our experiments. The low percentage of trainable tokens (only 5.12\%) highlights the potential memory savings that can be achieved through our logits masking technique, as described in Sec.~\ref{ssec:logits-masking}.

\begin{table}[htb]
\centering
\caption{Statistics of the XSum and SQuAD datasets used in our experiments}
\label{tab:xsum-squad-stats}
\begin{tabular}{l r r}
\toprule
\multirow{2}{*}{\textbf{Statistic (per sample)}} & \multicolumn{2}{c}{\textbf{Value [tokens]}} \\
 & XSum & SQuAD \\
\midrule
Avg. sequence length (w/o padding) & 531.47&238.76 \\
Avg. \# of trainable tokens &27.22 &5.41 \\
Avg. \% of trainable tokens & 5.12\%&2.40\% \\
\bottomrule
\end{tabular}
\end{table}
We used 50,000 training samples and 1,000 validation samples from the original dataset. To ensure that each sample would fit within the 2048 token context window, we had to remove a number of indices from the dataset. Specifically, we excluded approximately 1,000 samples with longer documents that would exceed our context length constraints when combined with the system prompt and tokenization overhead.

\subsubsection{SQuAD Dataset}
The SQuAD \cite{squad} dataset includes only training and validation splits. Following the approach used for XSum, we used 50,000 training samples and 1,000 validation samples from the original dataset. The chat template used for training is shown in Template~\ref{box:squad}.

Table~\ref{tab:xsum-squad-stats} presents key statistics about the SQuAD dataset used in our experiments. Similar to XSum, the low percentage of trainable tokens (only 2.4\%) highlights the potential memory savings that can be achieved through our logits masking technique.

\begin{myfloatbox2}[box:xsum]{XSum}
  \begin{tcolorbox}[system]
  You are a writing assistant that helps with summarizing text. You write summaries based on a user-provided text or article.
  \end{tcolorbox}

  \begin{tcolorbox}[user]
  Article: \verb|<$article>|
  \end{tcolorbox}

  \begin{tcolorbox}[assistant]
  Summary: \verb|<$summary>|
  \end{tcolorbox}

\end{myfloatbox2}
\begin{myfloatbox2}[box:squad]{SQuAD}
  \begin{tcolorbox}[system]
    You are a question answering assistant. You will be given a user-provided context (paragraph or sentence), followed by a question. Your job will be to answer the question based on the paragraph or sentence.
  \end{tcolorbox}

  \begin{tcolorbox}[user]
  Title: \verb|<$title>|
  
  Context: \verb|<$context>|
  
  Question: \verb|<$question>|
  \end{tcolorbox}

  \begin{tcolorbox}[assistant]
  Answer: \verb|<$answer>|
  \end{tcolorbox}

\end{myfloatbox2}

\clearpage
\section{Additional Experiments}

\subsection{Fine-tuning with SQuAD}
\label{app:SQuAD}

\begin{figure}[thb]
    \centering
    \includegraphics[width=0.9\textwidth]{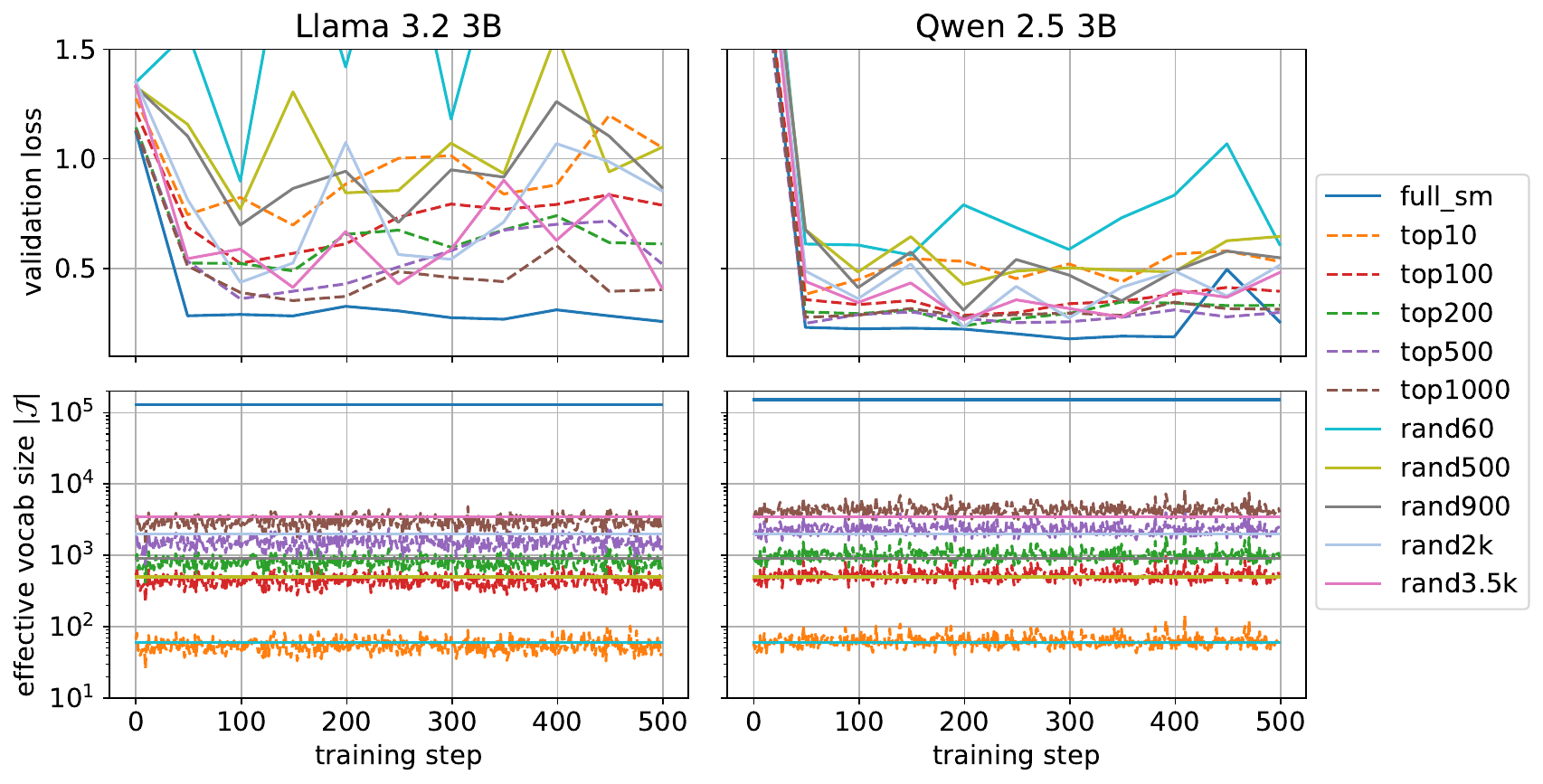} 
    \caption{Validation loss vs. training steps (top) and effective vocabulary size $|\mathcal{J}|$ (bottom) for \llama{} (left) and \qwen{} (right) fine-tuned on SQuAD using different softmax methods. Top-$k$ maintains convergence quality comparable to full softmax while using a fraction of the vocabulary.}
    \label{fig:squad}
\end{figure}

In Figure \ref{fig:squad}, we show the impact of softmax approximation on training convergence for the SQuAD dataset \cite{squad}. Note that in comparison to Figure \ref{fig:xsum_llama3} for the XSum dataset, the same choice of $k$ for top-k selection leads to a much smaller effective vocabulary size due to the inherently larger number of (trainable) tokens per data-point.

\clearpage
\subsection{Logits Masking w/o Softmax Approximation}\label{app:lm_no_sm}
In Sec.~\ref{ssec:logits-masking}, we reported results for logits masking when combined with quantization, checkpointing, and the top-1000 softmax approximation. To demonstrate that logits masking and softmax approximation are complementary, Table~\ref{tab:logits-mask-sm-0} presents results for logits masking with quantization and checkpointing only. Logits masking alone can significantly reduce peak memory for small percentages of trainable tokens, and combining logits masking with softmax approximation yields even greater savings.
\begin{table}[hb]
\centering
\caption{Impact of logits masking on peak memory vs. context length under different percentages of trainable tokens; we report results for checkpointing (full softmax). Colors indicate the memory bottleneck: \green{decoder} and \red{output}.}
\label{tab:logits-mask-sm-0}
\vskip 0.15in
\resizebox{0.99\textwidth}{!}{%
\begin{tabular}{c | r | r r r r  r r r r r r}
\toprule
\multirow{2.5}{*}{\textbf{ Model}} & \multirow{2.5}{*}{\textbf{Context Length}} &
\multicolumn{10}{c}{\textbf{Measured Peak Memory [GB]}} \\
\cmidrule(lr{1em}){3-12}
 & & 10\% & 20\% & 30\% & 40\% & 50\% & 60\% & 70\% & 80\% & 90\% & 100\% \\
\midrule
\multirow{7}{*}{\rotatebox{90}{\llama{}}}
 &512 & \red{1.88} & \red{1.91} & \red{1.94} & \red{1.96} & \red{1.99} & \red{2.01} & \red{2.04} & \red{2.06} & \red{2.09} & \red{2.11} \\
&1024 & \red{1.92} & \red{1.97} & \red{2.02} & \red{2.07} & \red{2.12} & \red{2.17} & \red{2.22} & \red{2.27} & \red{2.33} & \red{2.38} \\
&2048 & \red{1.98} & \red{2.08} & \red{2.18} & \red{2.29} & \red{2.39} & \red{2.49} & \red{2.59} & \red{2.77} & \red{3.06} & \red{3.36} \\
&4096 & \red{2.10} & \red{2.31} & \red{2.51} & \red{2.79} & \red{3.38} & \red{3.97} & \red{4.57} & \red{5.16} & \red{5.75} & \red{6.34} \\
&8192 & \green{3.38} & \green{3.38} & \red{4.02} & \red{5.20} & \red{6.39} & \red{7.57} & \red{8.76} & \red{9.94} & \red{11.12} & \red{12.31} \\
&12288 & \green{4.84} & \green{4.84} & \red{5.84} & \red{7.62} & \red{9.39} & \red{11.17} & \red{12.94} & \red{14.72} & \red{16.49} & \red{18.27} \\
&16384 & \green{6.29} & \green{6.29} & \red{7.66} & \red{10.03} & \red{12.40} & \red{14.77} & \red{17.13} & \red{19.50} & \red{21.87} & \red{24.24} \\
\midrule
\multirow{7}{*}{\rotatebox{90}{\qwen{}}}
 &512 & \red{1.50} & \red{1.53} & \red{1.56} & \red{1.59} & \red{1.62} & \red{1.65} & \red{1.68} & \red{1.71} & \red{1.74} & \red{1.77} \\
&1024 & \red{1.53} & \red{1.59} & \red{1.65} & \red{1.71} & \red{1.77} & \red{1.83} & \red{1.89} & \red{1.95} & \red{2.01} & \red{2.07} \\
&2048 & \red{1.60} & \red{1.72} & \red{1.84} & \red{1.96} & \red{2.08} & \red{2.41} & \red{2.76} & \red{3.11} & \red{3.46} & \red{3.81} \\
&4096 & \green{1.88} & \red{1.97} & \red{2.43} & \red{3.13} & \red{3.83} & \red{4.52} & \red{5.22} & \red{5.92} & \red{6.62} & \red{7.32} \\
&8192 & \green{3.36} & \green{3.36} & \red{4.55} & \red{5.95} & \red{7.35} & \red{8.74} & \red{10.14} & \red{11.54} & \red{12.94} & \red{14.33} \\
&12288 & \green{4.84} & \green{4.84} & \red{6.68} & \red{8.78} & \red{10.88} & \red{12.97} & \red{15.07} & \red{17.16} & \red{19.26} & \red{21.35} \\
&16384 & \green{6.32} & \green{6.32} & \red{8.81} & \red{11.60} & \red{14.40} & \red{17.19} & \red{19.98} & \red{22.78} & \red{25.57} & \red{28.37} \\
\bottomrule
\end{tabular}
}
\end{table}

\clearpage
\subsection{Comparison with Gradient Checkpointing}\label{app:grad-ckpt}

Table~\ref{tab:off-loading} compares peak memory usage with and without memory offloading across different context lengths. The baseline (w/o offloading) corresponds to gradient checkpointing as implemented in PyTorch~\cite{chen2016training}. Our offloading approach achieves substantial memory savings, especially for long contexts. For example, at 16K tokens, \llama{} reduces peak memory from 30.81\,GB to 24.05\,GB. At shorter lengths, the gains are also pronounced: for 512 tokens, \llama{} decreases from 3.77\,GB to 2.11\,GB. These results show that offloading complements checkpointing by further alleviating memory pressure.

\begin{table}[thb]
\centering
\caption{Peak memory usage during one checkpointed forward and backward pass for \llama{} and \qwen{} with and without offloading across varying context lengths.}
\label{tab:off-loading}
\vskip 0.15in
\begin{tabular}{l | l | r r r r r r r}
\toprule
\multirow{2.5}{*}{\textbf{Model}} & \multirow{2.5}{*}{\textbf{Offloading}} &
\multicolumn{7}{c}{\textbf{Measured Peak Memory [GB]}} \\
\cmidrule(lr{.5em}){3-9}
 & & 512 & 1024 & 2048 & 4096 & 8192 & 12288 & 16384 \\
\midrule
\multirow{2}{*}{\llama{}} & w/o & 3.77 & 4.20 & 5.51 & 9.11 & 16.35 & 23.57 & 30.81 \\
& w/ & 2.11 & 2.37 & 3.33 & 6.29 & 12.21 & 18.13 & 24.05 \\
\midrule
\multirow{2}{*}{\qwen{}} & w/o & 4.51 & 4.95 & 6.96 & 11.02 & 19.12 & 27.24 & 35.35 \\
& w/ & 1.76 & 2.06 & 3.79 & 7.28 & 14.27 & 21.26 & 28.24 \\
\bottomrule
\end{tabular}
\end{table}

\clearpage
\section{Additional Benchmarking on Llama-3.3-70B}\label{app:llama70b}
To evaluate the scalability of our techniques, we additionally profile Llama-3.3-70B. Tables~\ref{tab:sm_approx_llama3_70b} and~\ref{tab:logits_mask_llama3_70b} report peak memory under softmax approximation and logits masking, respectively. The same trends as for the 3B-class models hold: the LM-head bottleneck dominates at full softmax, and softmax approximation plus logits masking shifts the bottleneck back to the decoder layers.

\begin{table}[tbph]
\centering
\caption{Peak memory vs.\ context length under checkpointing and softmax approximation levels for Llama-3.3-70B. Colors indicate the memory bottleneck: \green{decoder}, \& \red{output}.}
\label{tab:sm_approx_llama3_70b}
\vskip 0.15in
\begin{tabular}{r | r r r r}
\toprule
\multirow{2.5}{*}{\textbf{Context Length}} & \multicolumn{4}{c}{\textbf{Measured Peak Memory [GB]}} \\
\cmidrule(lr{1em}){2-5}
& Full SM & Top-100 & Top-500 & Top-1000 \\
\midrule
32 & \red{4.93} & \green{1.72} & \green{1.71} & \green{1.71} \\
64 & \red{4.94} & \green{1.75} & \green{1.74} & \green{1.74} \\
128 & \red{4.98} & \green{1.81} & \green{1.81} & \green{1.81} \\
512 & \red{5.19} & \green{2.17} & \green{2.17} & \green{2.17} \\
1024 & \red{5.46} & \green{2.65} & \green{2.65} & \red{2.96} \\
2048 & \red{6.01} & \green{3.62} & \green{3.62} & \red{4.36} \\
4096 & \red{7.12} & \green{5.56} & \green{5.56} & \red{6.09} \\
8192 & \red{12.98} & \green{9.60} & \red{10.51} & \red{12.21} \\
12288 & \red{18.98} & \green{14.11} & \red{16.57} & \red{18.43} \\
\bottomrule
\end{tabular}
\end{table}

\begin{table}[!b]
\centering
\caption{Impact of logits masking on peak memory vs. context length under different percentages of trainable tokens for Llama-3.3-70B; we report results for checkpointing and top-1000 softmax approximation. Colors indicate the memory bottleneck: \green{decoder}, and \red{output}.}
\label{tab:logits_mask_llama3_70b}
\vskip 0.15in
\begin{tabular}{r | r r r r r r r r r r}
\toprule
\multirow{2.5}{*}{\textbf{Context Length}} & \multicolumn{10}{c}{\textbf{Measured Peak Memory [GB]}} \\
\cmidrule(lr{1em}){2-11}
& 10\% & 20\% & 30\% & 40\% & 50\% & 60\% & 70\% & 80\% & 90\% & 100\% \\
\midrule
32 & \green{1.71} & \green{1.71} & \green{1.71} & \green{1.71} & \green{1.71} & \green{1.71} & \green{1.71} & \green{1.71} & \green{1.71} & \green{1.71} \\
64 & \green{1.74} & \green{1.74} & \green{1.75} & \green{1.74} & \green{1.74} & \green{1.74} & \green{1.74} & \green{1.75} & \green{1.75} & \green{1.74} \\
128 & \green{1.81} & \green{1.81} & \green{1.81} & \green{1.81} & \green{1.81} & \green{1.81} & \green{1.81} & \green{1.81} & \green{1.81} & \green{1.81} \\
512 & \green{2.17} & \green{2.17} & \green{2.17} & \green{2.17} & \green{2.17} & \green{2.17} & \green{2.17} & \green{2.17} & \green{2.17} & \green{2.17} \\
1024 & \green{2.65} & \green{2.65} & \green{2.65} & \green{2.65} & \green{2.65} & \green{2.65} & \green{2.65} & \red{2.74} & \red{2.88} & \red{2.99} \\
2048 & \green{3.62} & \green{3.62} & \green{3.62} & \green{3.62} & \green{3.62} & \green{3.62} & \red{3.81} & \red{4.07} & \red{4.23} & \red{4.42} \\
4096 & \green{5.56} & \green{5.56} & \green{5.56} & \green{5.56} & \green{5.56} & \green{5.56} & \green{5.56} & \red{5.62} & \red{5.95} & \red{6.21} \\
8192 & \green{9.60} & \green{9.60} & \green{9.60} & \green{9.60} & \green{9.60} & \green{9.60} & \green{9.60} & \red{10.01} & \red{11.25} & \red{12.46} \\
12288 & \green{14.11} & \green{14.11} & \green{14.11} & \green{14.11} & \green{14.11} & \green{14.11} & \green{14.11} & \red{15.17} & \red{16.98} & \red{18.80} \\
\bottomrule
\end{tabular}
\end{table}
\clearpage
\section{Quantitative Fine-tuning Results}\label{app:quantitative-results}

We report fine-tuning convergence and task-specific quality metrics across six model configurations (Llama-3.2-3B in BF16 and our quantize/dequantize INT4 variant ``QDQ''; Qwen-2.5-3B in BF16 and QDQ; Qwen3-4B-Instruct; and Llama-3.3-70B in BF16) and four datasets (GSM8K, MATH, XSum, SQuAD). Evaluation losses are always computed using the full softmax cross-entropy, regardless of the loss used for training.

\begin{table*}[!htbp]
\subsection{GSM8K Results}\label{app:gsm8k-results}
\centering
\captionsetup{width=0.98\textwidth}
\small
\caption{Final evaluation loss for GSM8K experiments. Evaluation loss is always computed using the full Softmax CE.}
\begin{tabular}{lcccccc}
\toprule
Model & CE & TopK-1000 & TopK-500 & TopK-200 & TopK-100 & TopK-50 \\
\midrule
Llama-3.2-3B-BF16 & 0.4883 & 0.4887 & 0.4895 & 0.4920 & 0.4988 & 0.5204 \\
Llama-3.2-3B-QDQ & 0.4929 & 0.4935 & 0.4949 & 0.4971 & 0.5038 & 0.5179 \\
Llama-3.3-70B-BF16 & 0.4772 & 0.4775 & 0.4775 & 0.4776 & 0.4777 & 0.4777 \\
Qwen2.5-3B-BF16 & 0.4332 & 0.4332 & 0.4335 & 0.4363 & 0.4387 & 0.4459 \\
Qwen2.5-3B-QDQ & 0.4238 & 0.4255 & 0.4283 & 0.4313 & 0.4341 & 0.4562 \\
Qwen3-4B-Instruct & 0.3899 & 0.3901 & 0.3908 & 0.3948 & 0.4019 & 0.4183 \\
\bottomrule
\end{tabular}

\end{table*}

\begin{table*}[!htbp]
\centering
\captionsetup{width=0.98\textwidth}
\small
\caption{Average training loss over last 10\% of training for GSM8K experiments. Different TopK choices influence loss-scaling.}
\begin{tabular}{lcccccc}
\toprule
Model & CE & TopK-1000 & TopK-500 & TopK-200 & TopK-100 & TopK-50 \\
\midrule
Llama-3.2-3B-BF16 & 0.3982 & 0.3936 & 0.3901 & 0.3820 & 0.3699 & 0.3485 \\
Llama-3.2-3B-QDQ & 0.4382 & 0.4336 & 0.4294 & 0.4213 & 0.4082 & 0.3876 \\
Llama-3.3-70B-BF16 & 0.2678 & 0.2677 & 0.2677 & 0.2678 & 0.2677 & 0.2677 \\
Qwen2.5-3B-BF16 & 0.2870 & 0.2832 & 0.2798 & 0.2710 & 0.2618 & 0.2473 \\
Qwen2.5-3B-QDQ & 0.3248 & 0.3225 & 0.3193 & 0.3069 & 0.2954 & 0.2789 \\
Qwen3-4B-Instruct & 0.3417 & 0.3375 & 0.3343 & 0.3235 & 0.3108 & 0.2926 \\
\bottomrule
\end{tabular}

\end{table*}

\begin{table*}[!htbp]
\centering
\captionsetup{width=0.98\textwidth}
\small
\caption{Accuracy for GSM8K experiments.}
\begin{tabular}{lccccccc}
\toprule
Model & Base & CE & TopK-1000 & TopK-500 & TopK-200 & TopK-100 & TopK-50 \\
\midrule
Llama-3.2-3B-BF16 & 0.7058 & 0.6376 & 0.6293 & 0.6361 & 0.6376 & 0.6414 & 0.6641 \\
Llama-3.2-3B-QDQ & 0.7096 & 0.6123 & 0.6206 & 0.6277 & 0.6255 & 0.6156 & 0.6346 \\
Llama-3.3-70B-BF16 & 0.8772 & 0.8825 & 0.8855 & 0.8848 & 0.8825 & 0.8870 & 0.8923 \\
Qwen2.5-3B-BF16 & 0.7642 & 0.6687 & 0.6801 & 0.6687 & 0.6384 & 0.6353 & 0.6406 \\
Qwen2.5-3B-QDQ & 0.4867 & 0.6262 & 0.6171 & 0.5845 & 0.6065 & 0.5861 & 0.5815 \\
Qwen3-4B-Instruct & 0.8787 & 0.8029 & 0.8059 & 0.8036 & 0.7945 & 0.7892 & 0.7953 \\
\bottomrule
\end{tabular}

\end{table*}

\begin{table*}[!tbp]
\centering
\captionsetup{width=0.98\textwidth}
\small
\caption{Average effective vocabulary size for GSM8K experiments. For CE, this equals the model vocabulary size.}
\begin{tabular}{lcccccc}
\toprule
Model & CE & TopK-1000 & TopK-500 & TopK-200 & TopK-100 & TopK-50 \\
\midrule
Llama-3.2-3B-BF16 & 128256 & 14406 & 8341 & 4667 & 2751 & 1545 \\
Llama-3.2-3B-QDQ & 128256 & 13026 & 7467 & 4041 & 2332 & 1322 \\
Llama-3.3-70B-BF16 & 128256 & -- & -- & -- & -- & -- \\
Qwen2.5-3B-BF16 & 151936 & 16553 & 9442 & 4359 & 2398 & 1282 \\
Qwen2.5-3B-QDQ & 151936 & 15537 & 8740 & 3920 & 2211 & 1205 \\
Qwen3-4B-Instruct & 151936 & 15021 & 8647 & 4045 & 2232 & 1205 \\
\bottomrule
\end{tabular}

\end{table*}


\begin{table*}[!tbp]
\subsection{MATH Results}\label{app:math-results}
\centering
\captionsetup{width=0.98\textwidth}
\small
\caption{Final evaluation loss for MATH experiments. Evaluation loss is always computed using the full Softmax CE.}
\begin{tabular}{lcccccc}
\toprule
Model & CE & TopK-1000 & TopK-500 & TopK-200 & TopK-100 & TopK-50 \\
\midrule
Llama-3.2-3B-BF16 & 0.6204 & 0.6223 & 0.6261 & 0.6374 & 0.6561 & 0.6204 \\
Llama-3.2-3B-QDQ & 0.6447 & 0.6476 & 0.6528 & -- & 0.6813 & 0.6457 \\
Llama-3.3-70B-BF16 & -- & 0.5515 & 0.5513 & -- & 0.5515 & 0.5514 \\
Qwen2.5-3B-BF16 & 0.5624 & 0.5629 & 0.5640 & 0.5692 & 0.5795 & 0.5953 \\
Qwen2.5-3B-QDQ & 0.6042 & 0.6018 & 0.6041 & 0.6153 & 0.6264 & 0.6490 \\
Qwen3-4B-Instruct & 0.5446 & 0.5451 & 0.5464 & 0.5525 & 0.5632 & 0.5843 \\
\bottomrule
\end{tabular}

\end{table*}

\begin{table*}[!tbp]
\centering
\captionsetup{width=0.98\textwidth}
\small
\caption{Average training loss over last 10\% of training for MATH experiments. Different TopK choices influence loss-scaling.}
\begin{tabular}{lcccccc}
\toprule
Model & CE & TopK-1000 & TopK-500 & TopK-200 & TopK-100 & TopK-50 \\
\midrule
Llama-3.2-3B-BF16 & 0.5595 & 0.5537 & 0.5481 & 0.5363 & 0.5211 & 0.5595 \\
Llama-3.2-3B-QDQ & 0.5387 & 0.5336 & 0.5289 & -- & 0.5021 & 0.5391 \\
Llama-3.3-70B-BF16 & -- & 0.5111 & 0.5110 & 0.5112 & 0.5110 & 0.5110 \\
Qwen2.5-3B-BF16 & 0.5624 & 0.5575 & 0.5517 & 0.5386 & 0.5215 & 0.5001 \\
Qwen2.5-3B-QDQ & 0.5731 & 0.5620 & 0.5539 & 0.5434 & 0.5272 & 0.5013 \\
Qwen3-4B-Instruct & 0.5506 & 0.5461 & 0.5416 & 0.5276 & 0.5095 & 0.4848 \\
\bottomrule
\end{tabular}

\end{table*}

\begin{table*}[!tbp]
\centering
\captionsetup{width=0.98\textwidth}
\small
\caption{Normalized Match for MATH experiments.}
\resizebox{0.98\textwidth}{!}{%
\begin{tabular}{lccccccc}
\toprule
Model & Base & CE & TopK-1000 & TopK-500 & TopK-200 & TopK-100 & TopK-50 \\
\midrule
Llama-3.2-3B-BF16 & 0.6185 & 0.3922 & 0.3987 & 0.3972 & 0.3890 & 0.3707 & 0.3924 \\
Llama-3.2-3B-QDQ & 0.5856 & 0.2991 & 0.3086 & 0.3123 & -- & 0.2809 & 0.3081 \\
Llama-3.3-70B-BF16 & 0.8105 & -- & 0.6650 & 0.6649 & 0.6603 & 0.6586 & 0.6644 \\
Qwen2.5-3B-BF16 & 0.6444 & 0.3654 & 0.3714 & 0.3776 & 0.3728 & 0.3715 & 0.3584 \\
Qwen2.5-3B-QDQ & 0.2686 & 0.3187 & 0.3173 & 0.3110 & 0.3011 & 0.3026 & 0.2928 \\
Qwen3-4B-Instruct & 0.8876 & 0.5155 & 0.5153 & 0.5211 & 0.5280 & 0.5302 & 0.5102 \\
\bottomrule
\end{tabular}
}

\end{table*}

\begin{table*}[!tbp]
\centering
\captionsetup{width=0.98\textwidth}
\small
\caption{Average effective vocabulary size for MATH experiments. For CE, this equals the model vocabulary size.}
\begin{tabular}{lcccccc}
\toprule
Model & CE & TopK-1000 & TopK-500 & TopK-200 & TopK-100 & TopK-50 \\
\midrule
Llama-3.2-3B-BF16 & 128256 & 14697 & 9078 & 5292 & 3248 & -- \\
Llama-3.2-3B-QDQ & 128256 & 14697 & 9078 & -- & 3248 & -- \\
Llama-3.3-70B-BF16 & -- & -- & -- & -- & -- & -- \\
Qwen2.5-3B-BF16 & 151936 & 33854 & 20331 & 9750 & 5514 & 3101 \\
Qwen2.5-3B-QDQ & 151936 & 41713 & 25622 & 12671 & 7205 & 4079 \\
Qwen3-4B-Instruct & 151936 & 20144 & 12344 & 5941 & 3344 & 1842 \\
\bottomrule
\end{tabular}

\end{table*}


\begin{table*}[!tbp]
\subsection{XSum Results}\label{app:xsum-results}
\centering
\captionsetup{width=0.98\textwidth}
\small
\caption{Final evaluation loss for XSum experiments. Evaluation loss is always computed using the full Softmax CE.}
\begin{tabular}{lcccccc}
\toprule
Model & CE & TopK-1000 & TopK-500 & TopK-200 & TopK-100 & TopK-50 \\
\midrule
Llama-3.2-3B-BF16 & 1.4141 & 1.4506 & 1.4885 & 1.5505 & 1.6354 & 1.7493 \\
Llama-3.2-3B-QDQ & 1.4636 & 1.4989 & 1.5349 & 1.5827 & 1.6717 & 1.7929 \\
Llama-3.3-70B-BF16 & 1.0113 & 1.0191 & 1.0248 & 1.0479 & 1.0816 & 1.1254 \\
Qwen2.5-3B-BF16 & 1.3648 & 1.3863 & 1.4054 & 1.4582 & 1.5313 & 1.6378 \\
Qwen2.5-3B-QDQ & 1.4584 & 1.4857 & 1.5165 & 1.6322 & 1.7134 & 1.8120 \\
Qwen3-4B-Instruct & 1.4275 & 1.4547 & 1.4786 & 1.5423 & 1.6284 & 1.7440 \\
\bottomrule
\end{tabular}

\end{table*}

\begin{table*}[!tbp]
\centering
\captionsetup{width=0.98\textwidth}
\small
\caption{Average training loss over last 10\% of training for XSum experiments. Different TopK choices influence loss-scaling.}
\begin{tabular}{lcccccc}
\toprule
Model & CE & TopK-1000 & TopK-500 & TopK-200 & TopK-100 & TopK-50 \\
\midrule
Llama-3.2-3B-BF16 & 1.3857 & 1.2845 & 1.2251 & 1.1326 & 1.0457 & 0.9397 \\
Llama-3.2-3B-QDQ & 1.4349 & 1.3306 & 1.2701 & 1.1712 & 1.0841 & 0.9761 \\
Llama-3.3-70B-BF16 & 1.0048 & 0.9408 & 0.9074 & 0.8470 & 0.7897 & 0.7208 \\
Qwen2.5-3B-BF16 & 1.3313 & 1.2050 & 1.1500 & 1.0511 & 0.9634 & 0.8691 \\
Qwen2.5-3B-QDQ & 1.6624 & 1.2840 & 1.2240 & 1.3733 & 1.0326 & 0.9235 \\
Qwen3-4B-Instruct & 1.4095 & 1.2886 & 1.2284 & 1.1187 & 1.0148 & 0.9110 \\
\bottomrule
\end{tabular}

\end{table*}

\begin{table*}[!tbp]
\centering
\captionsetup{width=0.98\textwidth}
\small
\caption{ROUGE-1 F-measure for XSum experiments.}
\begin{tabular}{lccccccc}
\toprule
Model & Base & CE & TopK-1000 & TopK-500 & TopK-200 & TopK-100 & TopK-50 \\
\midrule
Llama-3.2-3B-BF16 & 0.2748 & 0.3885 & 0.3811 & 0.3768 & 0.3655 & 0.3518 & 0.3338 \\
Llama-3.2-3B-QDQ & 0.2678 & 0.3824 & 0.3724 & 0.3666 & 0.3599 & 0.3458 & 0.3265 \\
Llama-3.3-70B-BF16 & 0.2985 & 0.4672 & 0.4657 & 0.4650 & 0.4599 & 0.4528 & 0.4433 \\
Qwen2.5-3B-BF16 & 0.2478 & 0.3716 & 0.3651 & 0.3598 & 0.3493 & 0.3309 & 0.3082 \\
Qwen2.5-3B-QDQ & 0.2433 & 0.3783 & 0.3711 & 0.3634 & 0.3529 & 0.3360 & 0.3201 \\
Qwen3-4B-Instruct & 0.2485 & 0.3830 & 0.3798 & 0.3770 & 0.3679 & 0.3543 & 0.3369 \\
\bottomrule
\end{tabular}

\end{table*}

\begin{table*}[!tbp]
\centering
\captionsetup{width=0.98\textwidth}
\small
\caption{Average effective vocabulary size for XSum experiments. For CE, this equals the model vocabulary size.}
\begin{tabular}{lcccccc}
\toprule
Model & CE & TopK-1000 & TopK-500 & TopK-200 & TopK-100 & TopK-50 \\
\midrule
Llama-3.2-3B-BF16 & 128256 & 10538 & 5687 & 3355 & 2077 & 1206 \\
Llama-3.2-3B-QDQ & 128256 & 10538 & 5687 & 3355 & 2077 & 1206 \\
Llama-3.3-70B-BF16 & 128256 & 13193 & 7330 & 3570 & 1999 & 1110 \\
Qwen2.5-3B-BF16 & 151936 & 14762 & 8591 & 3854 & 2050 & 1072 \\
Qwen2.5-3B-QDQ & 151936 & 13901 & 7902 & 3612 & 1930 & 1010 \\
Qwen3-4B-Instruct & 151936 & 10675 & 6438 & 3103 & 1725 & 916 \\
\bottomrule
\end{tabular}

\end{table*}


\begin{table*}[!tbp]
\subsection{SQuAD Results}\label{app:squad-results}
\centering
\captionsetup{width=0.98\textwidth}
\small
\caption{Final evaluation loss for SQuAD experiments. Evaluation loss is always computed using the full Softmax CE.}
\begin{tabular}{lcccccc}
\toprule
Model & CE & TopK-1000 & TopK-500 & TopK-200 & TopK-100 & TopK-50 \\
\midrule
Llama-3.2-3B-BF16 & 0.1888 & 0.2004 & 0.2140 & 0.2535 & 0.2814 & 0.3019 \\
Llama-3.2-3B-QDQ & 0.1992 & 0.2072 & 0.2222 & 0.2598 & 0.3019 & 0.3220 \\
Llama-3.3-70B-BF16 & 0.1643 & 0.1815 & 0.1934 & 0.2120 & 0.2230 & 0.2398 \\
Qwen2.5-3B-BF16 & 0.1698 & 0.1952 & 0.1893 & 0.2054 & 0.2192 & 0.2368 \\
Qwen2.5-3B-QDQ & 0.1955 & 0.2236 & 0.2179 & 0.2876 & 0.2748 & 0.2856 \\
Qwen3-4B-Instruct & 0.1660 & 0.1741 & 0.1761 & 0.1805 & 0.1877 & 0.1950 \\
\bottomrule
\end{tabular}

\end{table*}

\begin{table*}[!tbp]
\centering
\captionsetup{width=0.98\textwidth}
\small
\caption{Average training loss over last 10\% of training for SQuAD experiments. Different TopK choices influence loss-scaling.}
\begin{tabular}{lcccccc}
\toprule
Model & CE & TopK-1000 & TopK-500 & TopK-200 & TopK-100 & TopK-50 \\
\midrule
Llama-3.2-3B-BF16 & 0.1667 & 0.1454 & 0.1392 & 0.1284 & 0.1188 & 0.1111 \\
Llama-3.2-3B-QDQ & 0.1860 & 0.1606 & 0.1507 & 0.1400 & 0.1277 & 0.1213 \\
Llama-3.3-70B-BF16 & 0.1555 & 0.1290 & 0.1218 & 0.1140 & 0.1087 & 0.1000 \\
Qwen2.5-3B-BF16 & 0.1602 & 0.1305 & 0.1236 & 0.1170 & 0.1097 & 0.1025 \\
Qwen2.5-3B-QDQ & 0.2235 & 0.1636 & 0.1516 & 0.1425 & 0.1544 & 0.1480 \\
Qwen3-4B-Instruct & 0.1517 & 0.1285 & 0.1252 & 0.1161 & 0.1109 & 0.1081 \\
\bottomrule
\end{tabular}

\end{table*}

\begin{table*}[!tbp]
\centering
\captionsetup{width=0.98\textwidth}
\small
\caption{F1 Score for SQuAD experiments.}
\begin{tabular}{lccccccc}
\toprule
Model & Base & CE & TopK-1000 & TopK-500 & TopK-200 & TopK-100 & TopK-50 \\
\midrule
Llama-3.2-3B-BF16 & 0.2772 & 0.8520 & 0.8478 & 0.8392 & 0.8205 & 0.8090 & 0.7912 \\
Llama-3.2-3B-QDQ & 0.2888 & 0.8439 & 0.8279 & 0.8098 & 0.8099 & 0.7875 & 0.7766 \\
Llama-3.3-70B-BF16 & 0.3077 & 0.8733 & 0.8637 & 0.8590 & 0.8520 & 0.8482 & 0.8435 \\
Qwen2.5-3B-BF16 & 0.2526 & 0.8445 & 0.8243 & 0.8279 & 0.8159 & 0.8038 & 0.7899 \\
Qwen2.5-3B-QDQ & 0.2605 & 0.8204 & 0.7898 & 0.8079 & 0.7611 & 0.7534 & 0.7704 \\
Qwen3-4B-Instruct & 0.2948 & 0.8561 & 0.8485 & 0.8477 & 0.8433 & 0.8373 & 0.8303 \\
\bottomrule
\end{tabular}

\end{table*}

\begin{table*}[!tbp]
\centering
\captionsetup{width=0.98\textwidth}
\small
\caption{Average effective vocabulary size for SQuAD experiments. For CE, this equals the model vocabulary size.}
\begin{tabular}{lcccccc}
\toprule
Model & CE & TopK-1000 & TopK-500 & TopK-200 & TopK-100 & TopK-50 \\
\midrule
Llama-3.2-3B-BF16 & 128256 & 3584 & 1861 & 1001 & 566 & 297 \\
Llama-3.2-3B-QDQ & 128256 & 3584 & 1861 & 1001 & 566 & 297 \\
Llama-3.3-70B-BF16 & 128256 & 4843 & 2601 & 1108 & 594 & 305 \\
Qwen2.5-3B-BF16 & 151936 & 4758 & 2500 & 1063 & 554 & 278 \\
Qwen2.5-3B-QDQ & 151936 & 5713 & 3071 & 1255 & 676 & 338 \\
Qwen3-4B-Instruct & 151936 & 4638 & 2462 & 1051 & 562 & 287 \\
\bottomrule
\end{tabular}

\end{table*}

\clearpage

\section{Softmax Approximation Gradient Fidelity}\label{app:sm-ablation}
\begin{figure}[htbp]
    \centering
    \includegraphics[width=0.8\linewidth]{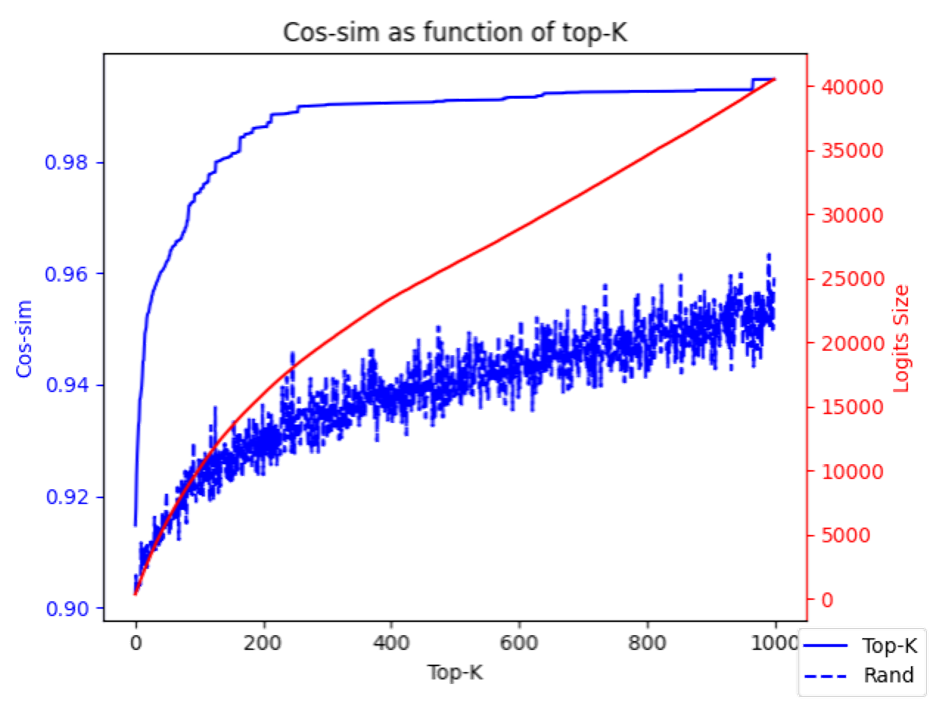}
    \caption{Cosine similarity between the exact gradient and the approximate gradient obtained using either top-$k$ or random sub-sampling of the output weight matrix $\vc{W}$ while sweeping $k$, showing that semantically based top-$k$ selection yields higher gradient fidelity with rapidly diminishing returns.}
    \label{fig:sm-gradient-fidelity}
\end{figure}

\section{Additional Dataset and Model Configurations}\label{app:extra-configs}

We complement the XSum and SQuAD dataset descriptions in Appendix~\ref{app:dataset} with details for the additional datasets used in our cross-dataset evaluation, as well as the model-specific configurations.

\subsection{GSM8K Dataset}\label{app:gsm8k}

The GSM8K dataset~\cite{cobbe2021gsm8k} consists of grade school math word problems requiring multi-step reasoning. We train models in chain-of-thought mode, where the model learns to generate both the reasoning process and the final numerical answer.

\subsubsection{Dataset Configuration}
We utilize the full GSM8K dataset from HuggingFace (\texttt{openai/gsm8k}), which includes approximately 7,500 training examples and 1,300 test examples. Each sample is formatted using a system prompt that instructs the model to solve mathematical problems step-by-step, as shown in Template~\ref{box:gsm8k}.

\begin{myfloatbox2}[box:gsm8k]{GSM8K}
  \begin{tcolorbox}[system]
  You are a helpful math assistant. Solve the following problem step by step.
  \end{tcolorbox}

  \begin{tcolorbox}[user]
  \verb|<$question>|
  \end{tcolorbox}

  \begin{tcolorbox}[assistant]
  \verb|<$reasoning>|

  \#\#\#\# \verb|<$answer>|
  \end{tcolorbox}
\end{myfloatbox2}

\subsubsection{Training Approach}
In chain-of-thought mode, the model is trained on the complete reasoning trace followed by the final answer in the standard GSM8K format (\texttt{\{reasoning\}\textbackslash n\#\#\#\# \{answer\}}). The system and user prompt tokens are masked (label = -100), while the full assistant response including both reasoning and answer is used for loss computation. This approach encourages the model to learn the intermediate reasoning steps rather than just memorizing answer patterns.

\subsubsection{Evaluation}
We evaluate model performance using accuracy, where we extract the numerical answer following the \texttt{\#\#\#\#} marker and compare it to the ground truth. The extraction handles various formats including numbers with commas and different answer phrasings.

\subsection{MATH Dataset}\label{app:math}

The MATH dataset~\cite{hendrycks2021measuring} contains challenging competition mathematics problems spanning algebra, geometry, number theory, and other advanced topics. Problems require sophisticated mathematical reasoning and answers are provided in LaTeX format within \texttt{\textbackslash boxed\{\}} delimiters.

\subsubsection{Dataset Configuration}
We load the MATH dataset from HuggingFace (\texttt{qwedsacf/competition\_math}). Since the original dataset contains only a single split, we create train and test splits using \texttt{train\_test\_split} with a fixed random seed (42) for reproducibility, allocating 5,000 examples for the test set and using the remainder for training. The chat template is shown in Template~\ref{box:math}.

\begin{myfloatbox2}[box:math]{MATH}
  \begin{tcolorbox}[system]
  You are a helpful math assistant. Solve the following problem step by step and provide your final answer.
  \end{tcolorbox}

  \begin{tcolorbox}[user]
  \verb|<$problem>|
  \end{tcolorbox}

  \begin{tcolorbox}[assistant]
  \verb|<$solution>|

  \textbackslash boxed\{\verb|<$answer>|\}
  \end{tcolorbox}
\end{myfloatbox2}

\subsubsection{Training Approach}
Similar to GSM8K, we train in chain-of-thought mode where models learn to generate the complete solution including intermediate steps. The solution text naturally includes the final answer in \texttt{\textbackslash boxed\{\}} format. We mask the problem tokens and train only on the solution tokens, allowing the model to learn both the reasoning process and proper LaTeX formatting for mathematical expressions.

\subsubsection{Evaluation}
We evaluate using normalized match accuracy, which accounts for mathematical equivalence rather than exact string matching. Our evaluation pipeline extracts answers from \texttt{\textbackslash boxed\{\}} expressions, normalizes them (removing whitespace, articles, and standardizing LaTeX commands), and uses SymPy for symbolic comparison when possible. This approach correctly identifies mathematically equivalent answers such as \texttt{0.5} and \texttt{\textbackslash frac\{1\}\{2\}}, or \texttt{x+1} and \texttt{1+x}.

\subsection{Model-Specific Configurations}\label{app:model-configs}

We experiment with six model configurations spanning different architectures, sizes, and precision formats:

\subsubsection{Model Variants}
\begin{itemize}
    \item \textbf{Llama-3.2-3B-Instruct}: A 3-billion parameter model with vocabulary size 128,256. We experiment with both the standard FP32 version and a quantized QDQ (Quantize-Dequantize) variant. Despite the ``BF16'' naming in our experiment labels, the 3B models are trained in FP32 precision. We use ``BF16'' version hosted on vLLM for task-specific metrics involving text generation.

    \item \textbf{Llama-3.3-70B-Instruct}: A 70-billion parameter model with vocabulary size 128,256, trained in BF16 precision. Due to memory constraints, we use gradient checkpointing and automatic device mapping (\texttt{device\_map="auto"}) to distribute the model across available GPUs.

    \item \textbf{Qwen2.5-3B-Instruct}: A 3-billion parameter model with vocabulary size 151,936. Similar to Llama-3.2-3B, we experiment with both FP32 and QDQ variants. We use ``BF16'' version hosted on vLLM for task-specific metrics involving text generation.

    \item \textbf{Qwen3-4B-Instruct-2507}~\cite{qwen3technicalreport}: A 4-billion parameter model with vocabulary size 151,936, trained in BF16 precision.
\end{itemize}

\subsubsection{Training Configuration}
All models are fine-tuned using LoRA (Low-Rank Adaptation) with model-specific hyperparameters. Key configuration choices include:

\begin{itemize}
    \item \textbf{Batch sizes}: The 70B model uses a per-device batch size of 1 with gradient accumulation steps of 4 across two GPUs, resulting in an effective batch size of 8, while smaller models use gradient accumulation with batch sizes of 4-8 depending on memory constraints but an effective batch size of 8 in all cases.

    \item \textbf{Context length}: All experiments use a maximum sequence length of 2048 tokens, with right-side padding using the model's pad token.

    \item \textbf{Chat template formatting}: Each model uses its native tokenizer's chat template to format conversations. Llama models add an EOS token during tokenization, while Qwen models do not.

    \item \textbf{Label masking}: For all datasets, we mask system and user prompt tokens (setting labels to -100) and train only on assistant response tokens. This focused training objective helps models learn to generate appropriate responses without being influenced by the prompt structure.

    \item \textbf{Top-k indices}: For experiments using top-k softmax, we pre-compute cosine similarity-based token indices for each model to identify the most relevant vocabulary subset for efficient training.
\end{itemize}

\subsubsection{Other Hyperparameters}
Training duration varies by dataset complexity and size:
\begin{itemize}
    \item GSM8K: 1,000 steps
    \item MATH: 2,813 steps
    \item XSum: 1,000 steps
    \item SQuAD: 500 steps
\end{itemize}

Learning rates were chosen heuristically rather than through systematic optimization, and are specific to each model but independent of the dataset and loss function. Across all experiments, we use a fixed LoRA configuration: rank-16 adapters on the Q and V projections with $\alpha=16$ and no dropout.

\end{document}